\documentclass[letterpaper]{article} 
\usepackage{main}  
\usepackage{times}  
\usepackage{helvet}  
\usepackage{courier}  
\usepackage[hyphens]{url}  
\usepackage{graphicx} 
\usepackage{natbib}  
\usepackage{caption} 
\usepackage{algorithm}
\usepackage{algorithmic}

\usepackage{amsmath}
\usepackage{amssymb}

\usepackage{hyperref}
\hypersetup{
    hidelinks,
    colorlinks=true,
    linkcolor=blue,
    citecolor=blue,
    urlcolor=blue
}

\usepackage{booktabs}
\usepackage{multirow}
\usepackage{array}
\usepackage{makecell}
\newcolumntype{C}[1]{>{\centering\arraybackslash}m{#1}}

\usepackage{newfloat}
\usepackage{listings}
\DeclareCaptionStyle{ruled}{labelfont=normalfont,labelsep=colon,strut=off} 
\floatstyle{ruled}
\newfloat{listing}{tb}{lst}{}
\floatname{listing}{Listing}

\title{\raisebox{-0.2cm}{\includegraphics[width=0.8cm]{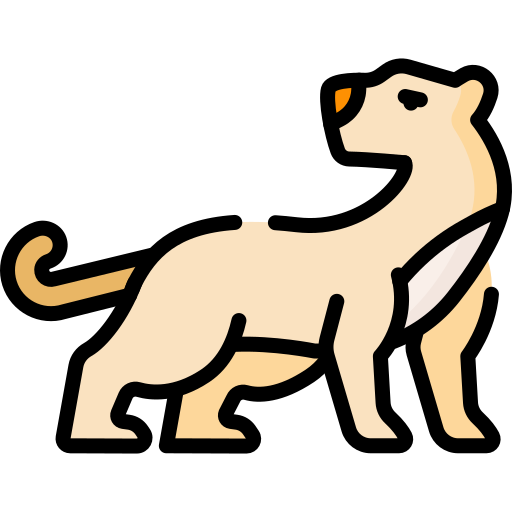}} Math-PUMA: Progressive Upward Multimodal Alignment to Enhance Mathematical Reasoning}
\author{
    Wenwen Zhuang\textsuperscript{\rm 1}\equalcontrib,
    Xin Huang\textsuperscript{\rm 2}\equalcontrib,
    Xiantao Zhang\textsuperscript{\rm 3}\equalcontrib,
    Jin Zeng\textsuperscript{\rm 1}
}
\affiliations{
    \textsuperscript{\rm 1}University of Chinese Academy of Sciences\\
    \textsuperscript{\rm 2}Beijing Institute of Technology
    \textsuperscript{\rm 3}Beihang University\\
    zhuangwenwen23@mails.ucas.ac.cn, huangxin@bit.edu.cn, zhangxiantao@buaa.edu.cn
}

\usepackage{bibentry}

\begin{document}
	
\maketitle

\begin{abstract}
Multimodal Large Language Models (MLLMs) excel in solving text-based mathematical problems, but they struggle with mathematical diagrams since they are primarily trained on natural scene images. For humans, visual aids generally enhance problem-solving, but MLLMs perform worse as information shifts from textual to visual modality. This decline is mainly due to their shortcomings in aligning images and text.
To tackle aforementioned challenges, we propose Math-PUMA, a methodology focused on \textbf{P}rogressive \textbf{U}pward \textbf{M}ultimodal \textbf{A}lignment. This approach is designed to improve the mathematical reasoning skills of MLLMs through a  three-stage training process, with the second stage being the critical alignment stage.
We first enhance the language model's mathematical reasoning capabilities with extensive set of textual mathematical problems. We then construct a multimodal dataset with varying degrees of textual and visual information, creating data pairs by presenting each problem in at least two forms. By leveraging the Kullback-Leibler (KL) divergence of next-token prediction distributions to align visual and textual modalities, consistent problem-solving abilities are ensured. Finally, we utilize multimodal instruction tuning for MLLMs with high-quality multimodal data. 
Experimental results on multiple mathematical reasoning benchmarks demonstrate that the MLLMs trained with Math-PUMA surpass most open-source MLLMs. Our approach effectively narrows the performance gap for problems presented in different modalities. The code and data are available at: \url{https://github.com/wwzhuang01/Math-PUMA}.
\end{abstract}

\section{Introduction}
Large Language Models (LLMs) have demonstrated remarkable reasoning capabilities, particularly when tackling mathematical problems in textual form \cite{wei2022chain,chen2022program,gou2023tora,yu2023metamath,shao2024deepseekmath}. However, Multimodal Large Language Models (MLLMs) face greater challenges when tackling problems that involve images. These models need to not only interpret textual information but also comprehend mathematical diagrams and identify details crucial for solving problems. Although MLLMs have exhibited notable efficacy in general visual question answering \cite{radford2021clip,li2022blip,liu2023llava}, their training predominantly relies on datasets comprising natural scene images. This reliance engenders a substantial domain discrepancy when these models are applied to mathematical diagrams, thereby resulting in inferior performance.

\begin{figure}[t!]
	\centering
	\begin{minipage}[t]{\linewidth}
		\centering
		\includegraphics[width=1.0\linewidth]{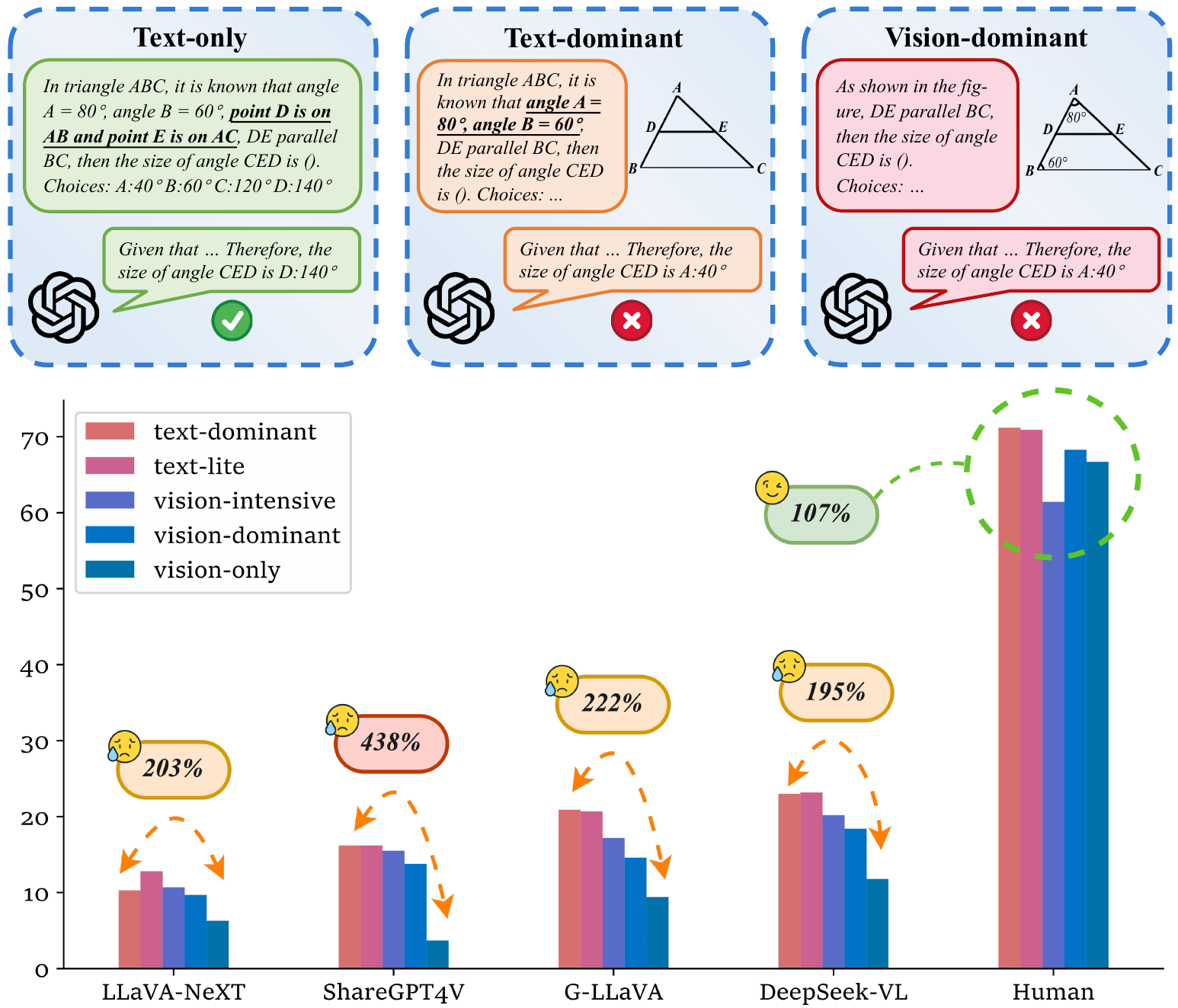}
		\caption{(\textit{Top}) Three examples of GPT-4o solving multimodal math problems. These examples represent different modalities of the same question. (\textit{Bottom}) Results of several open-source MLLMs and human on five different tasks of \textsc{MathVerse} \cite{zhang2024mathverse}.} \label{fig:mov}
	\end{minipage}
\end{figure}

For humans, regardless of the modality in which information is presented, problems with equivalent amounts of information tend to have similar levels of difficulty. Furthermore, incorporating images into problem-solving tasks can enhance human comprehension and resolution abilities. As illustrated in Figure \ref{fig:mov}, an increase in visual data often correlates with a decline in the efficacy of most MLLMs. Additionally, there is a notable disparity in effectiveness between text-centric and exclusively visual problems. For example, GPT-4o \cite{openai2024gpt4o} demonstrates strong proficiency in solving text-only mathematical problems, but its effectiveness diminishes progressively as the modality transitions from textual to visual. This reduction in capability primarily stems from the current models' inadequate alignment between visual and textual data, which impairs their overall functionality.

To address this issue, we propose Math-PUMA, a methodology centered around  \textbf{P}rogressive \textbf{U}pward \textbf{M}ultimodal \textbf{A}lignment (PUMA), aimed at enhancing the mathematical reasoning capabilities of MLLMs. Our approach is structured into three distinct stages, with stage 2 serving as the pivotal alignment phase.
\textbf{(1) Stage 1:}
We train the LLM using a substantial dataset of text-based math problems to enhance its problem-solving capabilities. This phase capitalizes on the extensive availability of text-based math problem-solving data.
\textbf{(2) Stage 2:}
It is observed that the model's mathematical problem-solving ability diminished progressively from text to vision, exhibiting an upward pyramidal structure. Consequently, the model's capabilities are categorized into four hierarchical levels. We construct 692K data pairs, with each pair conveying identical information but differing in multimodal representation. By leveraging the KL-divergence between next-token prediction distributions for text-rich and vision-rich problems, we achieve progressive bottom-up modal alignment across these hierarchical levels, thereby enhancing the model's ability to tackle multimodal mathematical problems.
\textbf{(3) Stage 3:}
We select 996K high-quality multimodal problem-solving data to fine-tune the model, further enhancing its performance in multimodal mathematical problem-solving tasks.

The contributions of this paper are three-fold:
\begin{itemize}
	\item We curate a large-scale dataset, Math-PUMA-1M, which comprises 692K data pairs and 996K multimodal mathematical data. This dataset serves as a valuable resource for model training.
	\item We propose Math-PUMA, a methodology based on Progressive Upward Multimodal Alignment, which enhances mathematical reasoning in MLLMs through a three-stage process.
	\item Experimental results on three widely-used benchmarks demonstrate that the MLLMs trained with Math-PUMA outperform most open-source models. Notably, our approach effectively narrows the performance gap for problems that contain the same information but are presented in different modalities, as evidenced by results on \textsc{MathVerse}.
\end{itemize}

\section{Related Work}
\subsection{Multimodal Large Language Models}
The exploration of Multimodal Large Language Models (MLLMs) has been inspired by advancements in Large Language Models (LLMs), resulting in remarkable capabilities across a variety of tasks that require both visual and linguistic understanding.
CLIP \cite{radford2021clip} is a breakthrough model that learns transferable visual representations from natural language supervision.
LLaVA series \cite{liu2023llava,liu2024improved}  pioneer visual instruction tuning for LLMs, employing a simple MLP as a projector to connect the vision encoder with the language model.
Models such as Qwen-VL \cite{bai2023qwen} and Deepseek-VL \cite{lu2024deepseek} introduce a new visual receptor or a hybrid vision encoder, significantly enhancing their ability to perceive and understand visual inputs.
However, despite these significant strides, MLLMs still face considerable challenges, particularly in multimodal mathematical reasoning. This is primarily due to the substantial domain gap between the natural scene image and the abstract  mathematical graphics. 
There is a pressing need to enhance the understanding and reasoning abilities of MLLMs in relation to mathematical diagrams.

\subsection{Multimodal Mathematical Reasoning}
The advancement of MLLMs has driven significant research into multimodal reasoning. Current efforts are primarily centered on data augmentation to improve models' performance. Significant efforts have been invested in augmenting text-only mathematical problem-solving data to enhance LLMs' reasoning capabilities \cite{saxton2019analysing,yu2023metamath,liu2024augmenting}. 
G-LLaVA \cite{gllava} and Math-LLaVA \cite{shi2024math} improve multimodal mathematical reasoning by constructing the Geo170K and MathV360K datasets, respectively.
These are created by generating additional questions for images sourced from public datasets.
However, they only serve to expand the text, without increasing the diversity of images in the dataset.
GeoGPT4V \cite{cai2024geogpt4v} leverages GPT-4V \cite{openai2023gpt4v} to generate new problems and images based on existing datasets, creating a dataset of 4.9K geometric problems combined with 19K open-source data. Nevertheless, due to GPT-4V's subpar capability in generating code from image descriptions, the quality of the generated data is comparatively inferior.
By comparison, our work not only makes new advancements in data augmentation, including text rephrasing and the generation of high-quality images, but also introduces a novel alignment method used for training.

\section{Methodology}
\begin{figure}[t!]
	\centering
	\begin{minipage}[t]{\linewidth}
		\centering
		\includegraphics[width=1.0\linewidth]{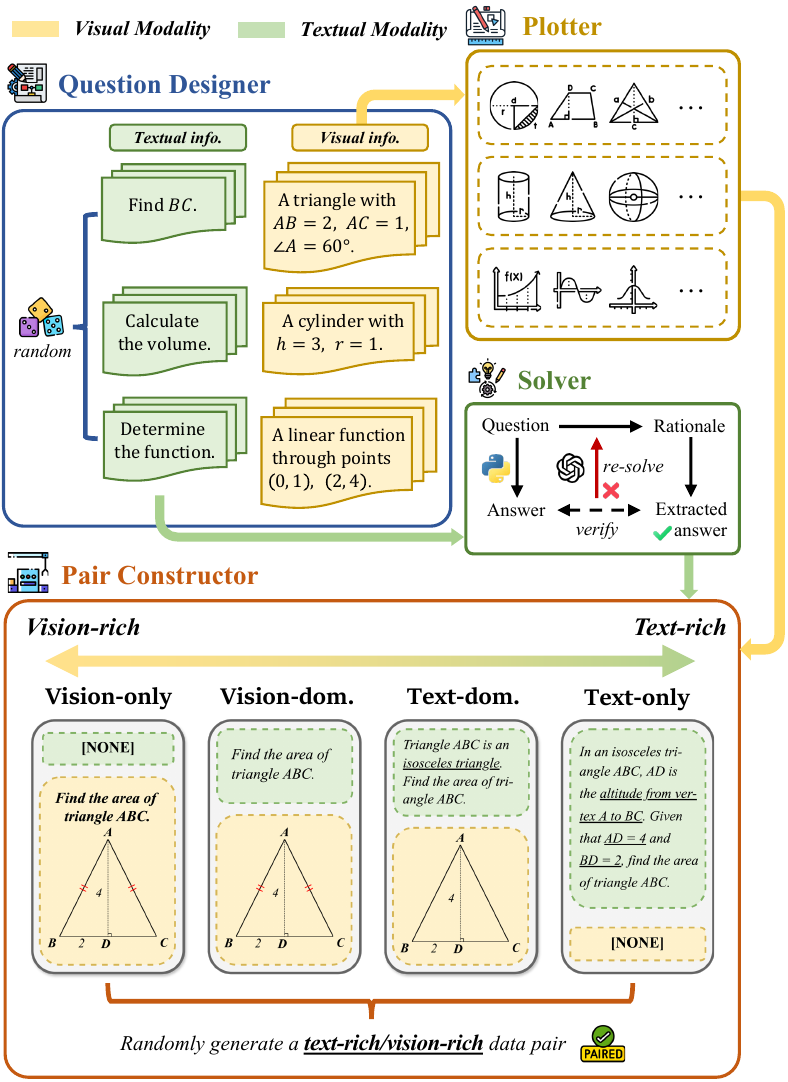}
		\caption{The pipeline of automatic data generation.} \label{fig:data-pipeline}
	\end{minipage}
\end{figure}
\begin{figure*}[ht!]
	\centering
	\begin{minipage}[t]{\linewidth}
		\centering
		\includegraphics[width=1.0\linewidth]{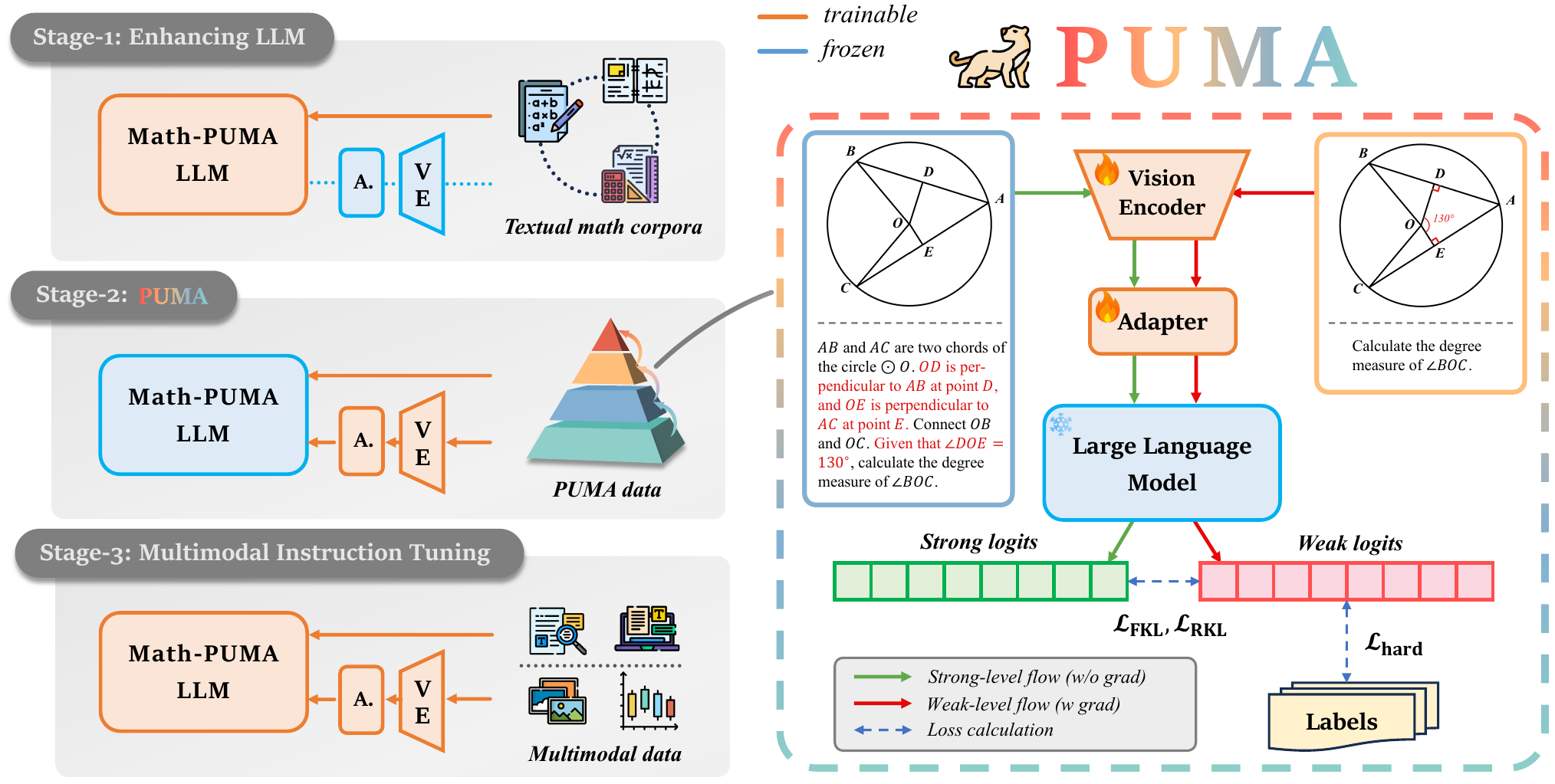}
		\caption{Overview of the Math-PUMA approach. (\textit{left}) The three stage training process of Math-PUMA. (\textit{right}) The details for aligning data pair. The input data pair includes text-rich data at the strong level and vision-rich data at the weak level, simultaneously processed by the MLLM. The strong logits and labels are used to supervise the weak logits.} \label{fig:puma-pyramid}
	\end{minipage}
\end{figure*}

\subsection{Data Construction}\label{sec:data-construction}
In order to refine alignment between visual and textual modalities, we need to construct data pairs. We clarify a \textbf{``data pair"} as a set of two data components, which share an equivalent level of information within each problem context, and their solutions are identical. A \textbf{``data pair"} is defined as two data components that contain equivalent information within the same problem context, and their solutions are identical. However, the distribution of information across different modalities may vary within each pair.
We use the term \textbf{``vision-rich"} to describe the data component where the visual modality has a higher proportion of information, whereas \textbf{``text-rich"} refers to the component with a higher proportion of textual information.

The methods we employ to construct data pairs include automatic data generation and data augmentation based on publicly available sources.

\subsubsection{Automatic Data Generation}
We implement an automatic data generation pipeline for three categories of mathematical problems: plane geometry, solid geometry, and functions. The pipeline consists of four agents: \textbf{(1) Question Designer}, responsible for formulating problems and assigning information to visual and textual modalities; \textbf{(2) Plotter}, which generates diagrams; \textbf{(3) Solver}, which provides answers and explanations; and \textbf{(4) Pair Constructor}, which produces four types of data and randomly selects two to form a data pair. Figure \ref{fig:data-pipeline} illustrates this automatic data generation process.
\begin{itemize}
	\item \textbf{Question Design:} 
	The Question Designer employs a random selection process to determine the specific type of mathematical problem to be generated. It also randomly selects the information carrier, deciding whether to present the information as text or an image. This choice dictates the visual information sent to the Plotter and the textual information sent to the Solver.
	\item \textbf{Plotting:}
	In accordance with the visual information received, the Plotter uses the predefined fundamental tools to plot diagrams.
	\item \textbf{Problem Solving:}
	The Solver calculates the answer using the text-only version of the problem, which contains complete information. As the calculation is performed programmatically, the answer is reliable. Considering that MLLMs can obtain stronger reasoning abilities from step-by-step solutions, the Solver generates a detailed explanation for each problem by calling GPT-4o-mini \cite{openai2024gpt4omini} and verifying the explanations against the standard answer to ensure accuracy.
	\item \textbf{Pair Construction:}
	The Pair Constructor combines the diagram from the Plotter and the text from the Solver to obtain up to four types of data, each comprising the same information but presented in a different modality: vision-only, vision-dominant, text-dominant, and text-only. Two of these are randomly selected to form a data pair, with the component containing more visual information classified as vision-rich and the other as text-rich.
\end{itemize}

We generated 40K data each for plane geometry, solid geometry, and functions, summing up to 120K.

\subsubsection{Data Augmentation}
We initially collect 80K mathematical problem-solving data from online sources. By rephrasing the problems from multiple perspectives \cite{yu2023metamath} and applying a series of traditional image processing techniques such as scaling, stretching, and gamma transformation, we expand the dataset to 310K. Additionally, we utilize the VisualWebInstruct dataset \cite{visualwebinstruct_datasets} containing 262K data. To automate the construction of data pairs, we employ a straightforward text-to-image rendering process to convert the content from textual to visual form. The original data serve as the text-rich component, while the generated data form the vision-rich component. In total, we obtain 572K data pairs.

\subsection{Training Stages}
We employ a three-stage pipeline to train our models, with specific details shown in Figure \ref{fig:puma-pyramid}.

\subsubsection{Stage 1: Enhancing the Language Model's Mathematical Reasoning Abilities}
Given the abundance of unsupervised text-based mathematical training corpora and problem-solving data \cite{shao2024deepseekmath}, in comparison to the scarcity of high-quality multimodal mathematical problem-solving data, we initially train the LLM on a large corpus of text-based math problems to bolster its mathematical reasoning capabilities.
To leverage the strengths of existing LLMs that have demonstrated superior performance in mathematical problem-solving \cite{shao2024deepseekmath,yang2024qwen2}, we use them to initialize our MLLMs. Subsequently, we fine-tune the model using 200,000 data points extracted from various datasets \cite{yue2023mammoth,tong2024dart,mitra2024orca,numina_math_datasets}. This phase significantly enhances the LLM's mathematical reasoning abilities.
\subsubsection{Stage 2: Progressive Upward Multimodal Alignment}
We observe that the multimodal mathematical reasoning ability of MLLMs resembles a pyramid, with performance declining from bottom to top as the information shifts from text to visual modalities. In order to address the discrepancy in performance between text-rich and vision-rich mathematical reasoning, we propose PUMA (Progressive Upward Multimodal Alignment). The objective of PUMA is to facilitate the effective resolution of vision-rich mathematical problems by aligning the outputs of MLLMs with text-rich data, thereby enhancing their reasoning capabilities across different modalities.

Let $i=0,1,2,3$ represents the levels of capability for MLLMs, ranging from weak to strong (top-down). For a visual mathematical problem, the inference results of MLLMs are progressively inferior on the $i$-th level compared to the $(i+1)$-th level. We denote the response distribution (logits) obtained by MLLMs when processing the input of $i$-th level as $p_{i}$, while the response distribution (logits) obtained on the input of $(i+1)$-th level is denoted as $p_{i+1}$. The forward KL (FKL) divergence and reverse KL (RKL) divergence between these distributions are calculated, since they converge to the same objective after a sufficient number of epochs for MLLMs \cite{wu2024rethinking}.

Let $\mathbf{y}^{(i)}=\{y^{(i)}_t\}^T_{t=1}$ represent the response generated by MLLMs based on input $\mathbf{x}^{(i)}$. Here, $y^{(i)}_t \in \{Y^{(i)}_1, Y^{(i)}_2, ..., Y^{(i)}_V\}$, with $V$ representing the vocabulary size. $p_{i}$ and $p_{i+1}$ represent the distributions of weak and strong levels, $\textbf{z}^{(i)}=(z^{(i)}_1,z^{(i)}_2,...,z^{(i)}_V)$ and $\textbf{z}^{(i+1)}=(z^{(i+1)}_1,z^{(i+1)}_2,...,z^{(i+1)}_V)$ represent the logits of weak and strong levels, respectively. The FKL divergence and RKL divergence are computed as follows:
\begin{equation}
	\resizebox{0.9\linewidth}{!}
	{$
		\begin{split}
			\mathcal{L}_\text{FKL}&=\frac{1}{TV}\sum^{T}_{t=1} {\text{KL}\Big(p_i(y^{(i)}_t | \mathbf{y}^{(i)}_{<t}) || p_{i+1}(y^{(i+1)}_t | \mathbf{y}^{(i+1)}_{<t})\Big)}\\
			&=\frac{1}{TV}\sum^{T}_{t=1}\sum^{V}_{j=1}p_i(Y^{(i)}_j | \mathbf{y}^{(i)}_{<t})\log\frac{p_i(Y^{(i)}_j | \mathbf{y}^{(i)}_{<t})}{p_{i+1}(Y^{(i+1)}_j | \mathbf{y}^{(i+1)}_{<t})},
		\end{split}
		$}
\end{equation}
\begin{equation}
	\resizebox{0.9\linewidth}{!}
	{$
		\begin{split}
			\mathcal{L}_\text{RKL}&=\frac{1}{TV}\sum^{T}_{t=1} {\text{KL}\Big(p_{i+1}(y^{(i+1)}_t | \mathbf{y}^{(i+1)}_{<t}) || p_{i}(y^{(i)}_t | \mathbf{y}^{(i)}_{<t})\Big)}\\
			&=\frac{1}{TV}\sum^{T}_{t=1}\sum^{V}_{j=1}p_{i+1}(Y^{(i+1)}_j | \mathbf{y}^{(i+1)}_{<t})\log\frac{p_{i+1}(Y^{(i+1)}_j | \mathbf{y}^{(i+1)}_{<t})}{p_i(Y^{(i)}_j | \mathbf{y}^{(i)}_{<t})},
		\end{split}
		$}
\end{equation}
with
\begin{equation}
	p_i(Y^{(i)}_j | \mathbf{y}^{(i)}_{<t})=\frac{\exp{(z_j^{(i)}/\tau)}}{\sum^V_{k=1}\exp({z_k^{(i)}/\tau})},
\end{equation}
where $\tau$ represents the temperature hyperparameter.

Furthermore, to maintain training stability, we calculate a hard loss by utilizing the solutions of mathematical problems as the ground truth labels, i.e.,
\begin{equation}
	\begin{split}
		\mathcal{L}_\text{hard} &= -\frac{1}{TV}\sum_{t=1}^T\log p_i(y_t^{(i)}|\mathbf{x}^{(i)},\mathbf{y}_{<t}^{(i)})\\
		&= -\frac{1}{TV}\sum_{t=1}^T\sum_{j=1}^V\log p_i(Y_j^{(i)}|\mathbf{x}^{(i)},\mathbf{y}_{<t}^{(i)}).
	\end{split}
\end{equation}
Finally, the total loss is computed as
\begin{equation}
	\resizebox{0.9\linewidth}{!}
	{$
		\mathcal{L}=\lambda_{\text{KL}} (\alpha_\text{KL} \mathcal{L}_\text{FKL}+(1-\alpha_\text{KL})\mathcal{L}_\text{RKL}){\tau}^2+(1-\lambda_\text{KL})\mathcal{L}_\text{hard},
		$}
\end{equation}

where \(\lambda_{\text{KL}}\) is a hyperparameter that balances the weight between the combined FKL and RKL divergences and the hard loss term, \(\alpha_\text{KL}\) is a weight hyperparameter that balances the contribution between \(\mathcal{L}_\text{FKL}\) and \(\mathcal{L}_\text{RKL}\). The purpose of multiplying KL by \(\tau^2\) is to equalize the gradients of the two losses.

At this stage, we use a total of 692K data pairs for training, which includes 120K data pairs automatically generated and 572K data pairs obtained through data augmentation based on publicly available data as described in Data Construction.

\subsubsection{Stage 3: Multimodal Instruction Tuning}
In the final phase, we enhance the model's reasoning capabilities by incorporating multimodal problem-solving data. Initially, we retain the majority of the high-quality data used in Stage 2 and augment our dataset with the MathV360K dataset \cite{shi2024math}. Specifically, we focus on enriching the geometric problem subset within MathV360K, expanding it from 40K to 120K in order to address the scarcity of geometric data. Furthermore, we integrated a balanced amount of textual data to prevent any modal imbalance between text and visual modalities. All data included detailed reasoning processes to guide the model's understanding and learning.

Ultimately, we compiled a large-scale instruction tuning dataset, comprising a total of 996K data points. This multimodal instruction tuning not only bolsters the model's reasoning and problem-solving abilities but also ensures that it can effectively leverage both textual and visual information for improved performance in mathematical problem-solving.

\section{Experiments}
\subsection{Experimental Setup}
\subsubsection{Models}
We validate the effectiveness of our method across various base models and scales, including DeepSeek-Math-7B \cite{shao2024deepseekmath}, Qwen2-1.5B and Qwen2-7B\cite{yang2024qwen2}, chosen as the LLM for Math-PUMA. To ensure the compatibility with DeepSeek-Math and DeepSeek-VL \cite{lu2024deepseek}, we adhere to the architecture of DeepSeek-VL. For Qwen2, we adopt a similar architecture to LLaVA, with the visual encoder designated as SigLIP-so400m-patch14-384 \cite{zhai2023sigmoid}.

\subsubsection{Benchmarks}
We conduct extensive experiments on three popular multimodal mathematical problem-solving benchmarks: \textsc{MathVerse} \cite{zhang2024mathverse}, \textsc{MathVista} \cite{lumathvista}, and \textsc{We-Math} \cite{qiao2024we}. \textsc{MathVerse} evaluates the multimodal mathematical reasoning abilities of MLLMs under five different conditions. \textsc{MathVista} comprises samples that require fine-grained, in-depth visual understanding and compositional reasoning, posing a challenge for all baseline models on this benchmark. \textsc{We-Math} is the first benchmark specifically designed to explore the problem-solving principles beyond the end-to-end performance.

\subsubsection{Evaluation and Metrics}
We refer to the leaderboards and adopt the official implementations of \textsc{MathVerse}, \textsc{MathVista}, and \textsc{We-Math}. For \textsc{MathVerse} and \textsc{MathVista}, initially, we use GPT-4o-mini \cite{openai2024gpt4omini} to extract answers from the responses generated by MLLMs. Subsequently, we employ GPT-4o-mini once more to verify the correctness of the extracted answers. The prompts used for answer extraction and correctness assessment are kept consistent with the official implementation. Ultimately, we calculate the accuracy scores as the evaluation metric. For \textsc{We-Math}, we select the average and Rote Memorization (RM) scores as evaluation metrics.

\subsubsection{Implementation Details}
Our experiments are conducted using PyTorch version 2.1.0 and CUDA 12.1, utilizing 32 NVIDIA A100 GPUs with 80GB memory each. The training process is divided into three stages, each with specific hyperparameters and configurations.
We employ the AdamW optimizer \cite{kingma2014adam}, configured with \(\beta_1 = 0.9\) and \(\beta_2 = 0.999\). The learning rate is adjusted across three stages: \(3 \times 10^{-5}\) for stage 1, \(5 \times 10^{-5}\) for stage 2, and \(3 \times 10^{-5}\) for stage 3. A cosine learning rate schedule is implemented with a warm-up phase covering 2\% of the total training steps. Additionally, a decay rate of 0.1 is applied.
The KL divergence is controlled using specific hyperparameters: \(\alpha_\text{KL}\) is set to 0.2,  \(\tau\) to 1.0, and \(\lambda_\text{KL}\) to 0.1.
The training is conducted over 1 epoch. The batch sizes for three stages are 256, 512, and 256, respectively.

\begin{table*}[!t] \centering
	\caption{\textbf{Mathematical evaluation on \textsc{MathVerse} and \textsc{MathVista} \textit{testmini} sets.} For \textsc{MathVerse}, we calculate the ``ALL" score without averaging the ``Text-only" version. For \textsc{MathVista}, we select 4 mathematical categories from the original 12 categories. ALL: overall accuracy across original categories; GPS: geometry problem solving; ALG: algebraic reasoning; GEO: geometry reasoning; SCI: scientific reasoning. For closed-source and open-source MLLMs, the best accuracy scores are marked in \textbf{bold} fonts, while the second best accuracy scores are marked in \underline{underline} fonts, respectively.}
	\resizebox{1.0\linewidth}{!}
	{
		\begin{tabular}{l|c|C{0.9cm}|C{0.9cm}|C{0.9cm}|C{0.9cm}|C{0.9cm}|C{0.9cm}|C{0.9cm}|C{0.9cm}|C{0.9cm}|C{0.9cm}|C{0.9cm}}
			\toprule
			\multirow{3}*{\large Model} & \multirow{3}*{\large \# Params.} & \multicolumn{6}{c|}{\textsc{MathVerse}} & \multicolumn{5}{c}{\textsc{MathVista}}\\
			\cmidrule{3-13}
			& &\multicolumn{1}{c|}{\makecell*[c]{ALL $\uparrow$}}
			&\multicolumn{1}{c|}{\makecell*[c]{\shortstack{Text-\\dom. $\uparrow$}}} 
			&\multicolumn{1}{c|}{\makecell*[c]{\shortstack{Text-\\lite $\uparrow$}}}
			&\multicolumn{1}{c|}{\makecell*[c]{\shortstack{Vision-\\int. $\uparrow$}}}
			&\multicolumn{1}{c|}{\makecell*[c]{\shortstack{Vision-\\dom. $\uparrow$}}}
			&\multicolumn{1}{c|}{\makecell*[c]{\shortstack{Vision-\\only $\uparrow$}}}
			&\multicolumn{1}{c|}{\makecell*[c]{\shortstack{ALL $\uparrow$}}}
			&\multicolumn{1}{c|}{\makecell*[c]{\shortstack{GPS $\uparrow$}}}
			&\multicolumn{1}{c|}{\makecell*[c]{\shortstack{ALG $\uparrow$}}}
			&\multicolumn{1}{c|}{\makecell*[c]{\shortstack{GEO $\uparrow$}}}
			&\multicolumn{1}{c}{\makecell*[c]{\shortstack{SCI $\uparrow$}}}
			\\
			\midrule
			\multicolumn{13}{c}{\textit{Baselines}}\\
			\cmidrule{1-13}
			Random chance & - & 12.4 & 12.4 & 12.4 & 12.4 & 12.4 & 12.4 & 17.9 & 21.6 & 21.7 & 20.1 & 17.2 \\
			Human performance & - & 64.9 & 71.2 & 70.9 & 61.4 & 68.3 & 66.7 & 60.3 & 48.4 & 50.9 & 51.4 & 64.9 \\
			\cmidrule{1-13}
			\multicolumn{13}{c}{\textit{Closed-source LLMs}}\\
			\cmidrule{1-13}
			ChatGPT~\cite{ouyang2022training} & - & - & 33.3 & 18.9 & - & - & - & 33.2 & 29.3 & 31.0 & 31.0 & 50.8 \\
			GPT-4~\cite{OpenAI2023GPT4TR} & - & - & 46.5 & 20.7 & - & - & - & 33.2 & 31.7 & 33.5 & 32.2 & 58.2 \\
			\cmidrule{1-13}
			\multicolumn{13}{c}{\textit{Closed-source MLLMs}}\\
			\cmidrule{1-13}
			Qwen-VL-Plus~\cite{bai2023qwen} & - & 11.8 & 15.7 & 11.1 & 9.0 & 13.0 & 10.0 & 43.3 & 38.5 & 39.1 & 39.3 & 59.0 \\
			Gemini-1.0-Pro~\cite{team2023gemini} & - & 22.3 & 27.6 & 23.7 & 19.4 & 20.3 & 20.5 & 45.2 & 40.4 & 45.2 & 41.0 & 54.9 \\
			Qwen-VL-Max~\cite{bai2023qwen} & - & 24.8 & 30.3 & 24.8 & 20.6 & 23.3 & 25.1 & - & - & - & - & - \\
			GPT-4V~\cite{openai2023gpt4v} & - & \textbf{38.3} & \textbf{52.1} & \textbf{40.9} & \textbf{34.9} & \textbf{33.6} & \textbf{29.8} & \textbf{49.9} & \textbf{50.5} & \textbf{53.0} & \textbf{51.0} & \textbf{63.1} \\
			\cmidrule{1-13}
			\multicolumn{13}{c}{\textit{Open-source MLLMs}}\\
			\cmidrule{1-13}
			mPLUG-Owl2~\cite{ye2024mplug} & 7B & 4.6 & 6.6 & 6.3 & 6.3 & 5.6 & 4.9 & 22.2 & 23.6 & 23.6 & 23.9 & 26.3 \\
			LLaMA-Adapter-V2~\cite{gao2023llamaadapterv2} & 7B & 5.7 & 6.2 & 5.9 & 6.1 & 4.2 & 6.1 & 23.9 & 25.5 & 26.3 & 24.3 & 29.5 \\
			LLaVA-1.5~\cite{liu2024improved} & 13B & 7.6 & 8.8 & 7.6 & 7.4 & 7.4 & 6.9 & 25.7 & 18.3 & 19.6 & 17.6 & 42.6 \\
			LLaVA-NeXT~\cite{liu2024llavanext} & 8B & 10.3 & 12.8 & 12.0 & 10.7 & 9.7 & 6.3 & 34.6 & - & - & - & - \\
			MiniGPT-v2~\cite{chen2023minigpt} & 7B & 11.0 & 12.1 & 12.0 & 13.1 & 10.3 & 7.4 & 23.1 & 26.0 & 28.1 & 24.7 & 25.4 \\
			SPHINX-Plus~\cite{gao2024sphinx} & 13B & 12.2 & 13.9 & 11.6 & 11.6 & 13.5 & 10.4 & 36.8 & - & - & - & - \\
			ShareGPT4V~\cite{Chen2023ShareGPT4VIL} & 13B & 13.1 & 16.2 & 16.2 & 15.5 & 13.8 & 3.7 & 27.5 & 27.4 & - & 27.6 & - \\
			InternLM-XC2.~\cite{dong2024internlm} & 7B & 16.3 & 20.2 & 14.3 & 14.2 & 17.5 & 15.2 & \underline{47.8} & 31.7 & 32.0 & 30.5 & 37.7 \\
			G-LLaVA~\cite{gllava} & 7B & 16.6 & 20.9 & 20.7 & 17.2 & 14.6 & 9.4 & 23.8 & 38.9 & 36.3 & 35.6 & 20.5 \\
			SPHINX-MoE~\cite{gao2024sphinx} & 8$\times$7B & 16.8 & 26.2 & 17.4 & 16.7 & 12.5 & 11.1 & 42.3 & 31.2 & 31.7 & 30.5 & \textbf{50.8} \\
			DeepSeek-VL~\cite{lu2024deepseek} & 7B & 19.3 & 23.0 & 23.2 & 20.2 & 18.4 & 11.8 & 34.9 & 28.4 & 29.2 & 27.2 & 35.3 \\
			Math-LLaVA~\cite{shi2024math} & 13B & 22.9 & 27.3 & 24.9 & 24.5 & 21.7 & 16.1 & 38.3 & 29.3 & 28.5 & 30.5 & 42.6 \\
			\midrule
			Math-PUMA-Qwen2-1.5B & 1.5B & 29.6 & 35.8 & 32.2 & 31.3 & \underline{30.4} & \underline{18.5} & 44.5 & \underline{47.6} & \underline{43.4} & \textbf{47.3} & 41.0 \\
			Math-PUMA-Qwen2-7B & 7B & \textbf{33.6} & \underline{42.1} & \underline{35.0} & \underline{33.4} & \textbf{31.6} & \textbf{26.0} & \textbf{47.9} & \textbf{48.1} & \textbf{47.7} & \textbf{47.3} & 42.6 \\
			Math-PUMA-DeepSeek-Math-7B & 7B & \underline{31.8} & \textbf{43.4} & \textbf{35.4} & \textbf{33.6} & \textbf{31.6} & 14.7 & 44.7 & 39.9 & 39.2 & \underline{41.4} & \underline{48.4} \\
			\bottomrule
		\end{tabular}
	}
	\label{tab:mathv}
\end{table*}

\begin{table}[t!] \centering
	\caption{\textbf{Evaluation results on \textsc{We-Math} \textit{testmini} set.} AVG: average score (strict); RM: rote memorization (strict). The best scores of each category are marked in \textbf{bold} fonts.}
	\setlength{\tabcolsep}{1pt}
	\resizebox{0.95\linewidth}{!}{
		\begin{tabular}{l|c|>{\large}C{1.4cm}|>{\large}C{1.4cm}}
			\toprule
			Model & \# Params. & AVG $\uparrow$ & RM $\downarrow$ \\ 
			\cmidrule{1-4}
			\multicolumn{4}{c}{\textit{Closed-source MLLMs}}\\
			\cmidrule{1-4}        
			Qwen-VL-Max~\cite{bai2023qwen} & - & 10.5 & 75.5 \\
			Gemini-1.5-Pro~\cite{reid2024gemini} & - & 26.4 & 54.8 \\
			GPT-4V~\cite{openai2023gpt4v} & - & 31.1 & 47.9 \\
			GPT-4o~\cite{openai2024gpt4o} & - & \textbf{42.9} & \textbf{34.2} \\
			\cmidrule{1-4}
			\multicolumn{4}{c}{\textit{Open-source MLLMs ($\geq$20B)}}\\
			\cmidrule{1-4}
			InternVL-Chat-V1.5~\cite{chen2024far} & 26B & 15.0 & 73.3 \\
			LLaVA-NeXT~\cite{liu2024llavanext} & 72B & 13.4 & 71.0 \\
			LLaVA-NeXT~\cite{liu2024llavanext} & 110B & \textbf{19.2} & \textbf{66.0} \\
			\cmidrule{1-4}
			\multicolumn{4}{c}{\textit{Open-source MLLMs ($\approx$10B)}}\\
			\cmidrule{1-4}
			LLaVA-1.5~\cite{liu2024improved} & 7B & 6.5 & 85.6 \\
			LLaVA-1.5~\cite{liu2024improved} & 13B & 8.4 & 78.1 \\
			LLaVA-1.6~\cite{liu2024llavanext} & 7B & 3.3 & 89.1 \\
			LLaVA-1.6~\cite{liu2024llavanext} & 13B & 5.2 & 86.9 \\
			DeepSeek-VL~\cite{lu2024deepseek} & 7B & 6.3 & 84.8 \\
			G-LLaVA~\cite{gllava} & 13B & 6.5 & 86.6 \\
			Math-LLaVA~\cite{shi2024math} & 13B & 11.1 & 72.8 \\
			InternLM-XC2.~\cite{dong2024internlm} & 7B & 12.7 & 77.6 \\
			\midrule
			Math-PUMA-Qwen2-1.5B & 1.5B & 10.4 & 75.5 \\
			Math-PUMA-Qwen2-7B & 7B & \textbf{19.2} & 67.8 \\
			Math-PUMA-DeepSeek-Math & 7B & 15.6 & \textbf{67.4} \\
			\bottomrule
		\end{tabular}
	}
	\label{tab:wemath}
\end{table}

\begin{table}[ht!] \centering
	\caption{\textbf{Results of ablation study.} Order: the sequential order of Stage 1, 2, and 3; ALL: overall accuracy; Text-dom.: accuracy of text-dominant data; Vision-only: accuracy of vision-only data; Gap: (Text-dom. $-$ Vision-only) / Vision-only. The best scores of each LLM are marked in \textbf{bold} fonts.}
	\setlength{\tabcolsep}{2pt}
	\resizebox{0.95\linewidth}{!}
	{
		\begin{tabular}{c|c|C{1.2cm}|C{1.2cm}|C{1.2cm}|C{1.2cm}}
			\toprule
			Order & LLM & \makecell*[c]{\shortstack{ALL $\uparrow$}} & \makecell*[c]{\shortstack{Text-\\dom. $\uparrow$}} & \makecell*[c]{\shortstack{Vision-\\only $\uparrow$}} & Gap $\downarrow$ \\ 
			\cmidrule{1-6}
			\multicolumn{6}{c}{\textit{Standard pipeline}}\\
			\cmidrule{1-6}
			\multirow{3}*{\makecell*[c]{\shortstack{1 $\rightarrow$ 2 $\rightarrow$ 3}}} 
			& Qwen2-1.5B & \textbf{29.6} & 35.8 & \textbf{18.5} & 93.5 \\
			& Qwen2-7B & \textbf{33.6} & 42.1 & \textbf{26.0} & \textbf{61.9} \\
			& DeepSeek-Math & \textbf{31.8} & \textbf{43.4} & 14.7 & 195.2 \\
			\cmidrule{1-6}
			\multicolumn{6}{c}{\textit{Effectiveness of Stage 1 (Enhancing LLM)}}\\
			\cmidrule{1-6}
			\multirow{3}*{2 $\rightarrow$ 3} 
			& Qwen2-1.5B & 17.0 & 19.9 & 12.1 & \textbf{64.5} \\
			& Qwen2-7B & 19.6 & 27.3 & 11.9 & 129.4 \\
			& DeepSeek-Math & 23.9 & 30.7 & 11.2 & 174.1 \\
			\cmidrule{1-6}
			\multicolumn{6}{c}{\textit{Effectiveness of Stage 2 (Math-PUMA)}}\\
			\cmidrule{1-6}
			\multirow{3}*{1 $\rightarrow$ 3} 
			& Qwen2-1.5B & 24.6 & \textbf{40.3} & 9.8 & 311.2 \\
			& Qwen2-7B & 27.2 & \textbf{44.1} & 11.0 & 300.9 \\
			& DeepSeek-Math & 29.3 & \textbf{43.4} & 9.1 & 376.9 \\
			\cmidrule{1-6}
			\multicolumn{6}{c}{\textit{Effectiveness of Stage 3 (Multimodal instruction tuning)}}\\
			\cmidrule{1-6}
			\multirow{3}*{1 $\rightarrow$ 2} 
			& Qwen2-1.5B & 11.7 & 15.5 & 8.1 & 91.4 \\
			& Qwen2-7B & 21.2 & 28.9 & 12.2 & 136.9 \\
			& DeepSeek-Math & 22.2 & 36.2 & \textbf{14.8} & \textbf{144.6} \\
			\cmidrule{1-6}
			\multicolumn{6}{c}{\textit{Sequential Order of Stages}}\\
			\cmidrule{1-6}
			\multirow{3}*{1 $\rightarrow$ 3 $\rightarrow$ 2} 
			& Qwen2-1.5B & 24.5 & 38.2 & 12.1 & 215.7 \\
			& Qwen2-7B & 26.7 & 34.4 & 18.7 & 84.0 \\
			& DeepSeek-Math & 23.4 & 34.3 & 4.3 & 697.7 \\
			\bottomrule
		\end{tabular}
	}
	\label{tab:ablation}
\end{table}

\subsection{Performance Comparison}
\subsubsection{Comparison on \textsc{MathVerse}}
\textsc{MathVerse} is capable of clearly demonstrating the gap between visual and textual modalities. From Table \ref{tab:mathv}, it can be observed that the MLLMs trained by Math-PUMA achieve the state-of-the-art (SOTA) among open-source MLLMs. Compared to the previous SOTA method, Math-LLaVA, the MLLMs trained by Math-PUMA  exhibit accuracy scores improvement about 10\%. When compared to the closed-source GPT-4V \cite{openai2023gpt4v}, Math-PUMA-Qwen2-7B performs competitively with only a gap of 4.7\%, demonstrating the effectiveness of Math-PUMA.

\subsubsection{Comparison on \textsc{MathVista}}
\textsc{MathVista} is a comprehensive benchmark designed to evaluate mathematical reasoning. According to the results presented in Table \ref{tab:mathv}, Math-PUMA-Qwen2-7B demonstrates SOTA performance in GPS, ALG, GEO and SCI domains among open-source MLLMs of the same scale. It outperforms InternLM-XComposer2-VL \cite{dong2024internlm} by significant margins, with accuracy improvements of 16.4\%, 15.7\%, 16.8\%, and 4.9\% in these respective domains.

\subsubsection{Comparison on \textsc{We-Math}}
\textsc{We-Math} places strong emphasis on the importance of the mathematical reasoning process. Table \ref{tab:wemath} demonstrates that Math-PUMA-Qwen2-7B achieves SOTA performance in average scores among open-source MLLMs with approximate 10B parameters, surpassing InternLM-XComposer2-VL. Notably, even among open-source MLLMs with parameters exceeding 20B, Math-PUMA-Qwen2-7B outperforms LLaVA-NeXT \cite{liu2024llavanext} 72B model, reaching the performance of LLaVA-NeXT 110B model. While Math-PUMA-Qwen2-7B surpasses Qwen-VL-Max \cite{bai2023qwen} among closed-source models, there remains a significant gap compared to GPT-4V and GPT-4o.

\subsection{Ablation Study}
Ablation studies are performed on \textsc{Mathverse} to highlight the contribution of each training stage and to assess the impact of their sequential order on Math-PUMA.

\subsubsection{The Role of Each Stage}
To evaluate the significance of each stage, we conduct three ablation experiments by individually removing stages 1, 2, and 3. We then assess the accuracy across overall, text-dominant, and vision-only scenarios, as well as the gaps between them. The results of these experiments are summarized in Table \ref{tab:ablation}.

\textbf{Removing Stage 1:} Stage 1 aims to enhance the mathematical reasoning capabilities of the LLMs. As observed in Table \ref{tab:ablation}, upon removing stage 1, there is a slight decrease in the accuracy compared to the corresponding model trained with all three stages. This reduction occurs because stage 1 serves as the foundation for stage 2. When the LLM lacks strong mathematical reasoning capabilities, strong logits are not reliable to supervise weak logits, resulting in lower performance. However, due to the presence of complete stage 2 and 3, the gap remains close to that of the complete three-stage training model and relatively low.

\textbf{Removing Stage 2:} Stage 2 embodies our devised PUMA, facilitating a close alignment between visual and textual modalities. As depicted in Table \ref{tab:ablation}, the absence of stage 2 results in a wider gap in reasoning performance between textual and visual modalities when compared to the three-stage approach. Nonetheless, with the enhancement of mathematical reasoning capabilities by stage 1 and multimodal instruction tuning with high-quality data through stage 3, the overall performance persists at a relatively high level.

\textbf{Removing Stage 3:} Stage 3 is multimodal instruction tuning. We observe that if only stage 1 and 2 are performed without subsequent multimodal instruction tuning, MLLMs tend to lose conversational capabilities to some extent. As seen in Table \ref{tab:ablation}, the performance of MLLMs drastically declines when stage 3 is excluded, primarily due to the loss of conversational capabilities. Since we have conducted stage 2, the gap between textual and visual modalities remains relatively small.

\subsubsection{Sequential Order of Stages}
We swap stage 2 and 3 to assess their impact on MLLMs. As shown in Table \ref{tab:ablation}, exchanging stage 2 and 3 leads to a significant performance drop. Our analysis of each stage reveals the critical role of stage 3 in maintaining the conversational abilities of MLLMs. Consequently, rearranging the stage 2 and 3 results in the loss of conversational skills of MLLMs, thereby influencing their overall performance. Nonetheless, the eventual implementation of stage 2 ensures that the gap between textual and visual modalities remains relatively small.
\begin{figure}[h!]
	\centering
	\begin{minipage}[t]{\linewidth}
		\centering
		\includegraphics[width=1.0\linewidth]{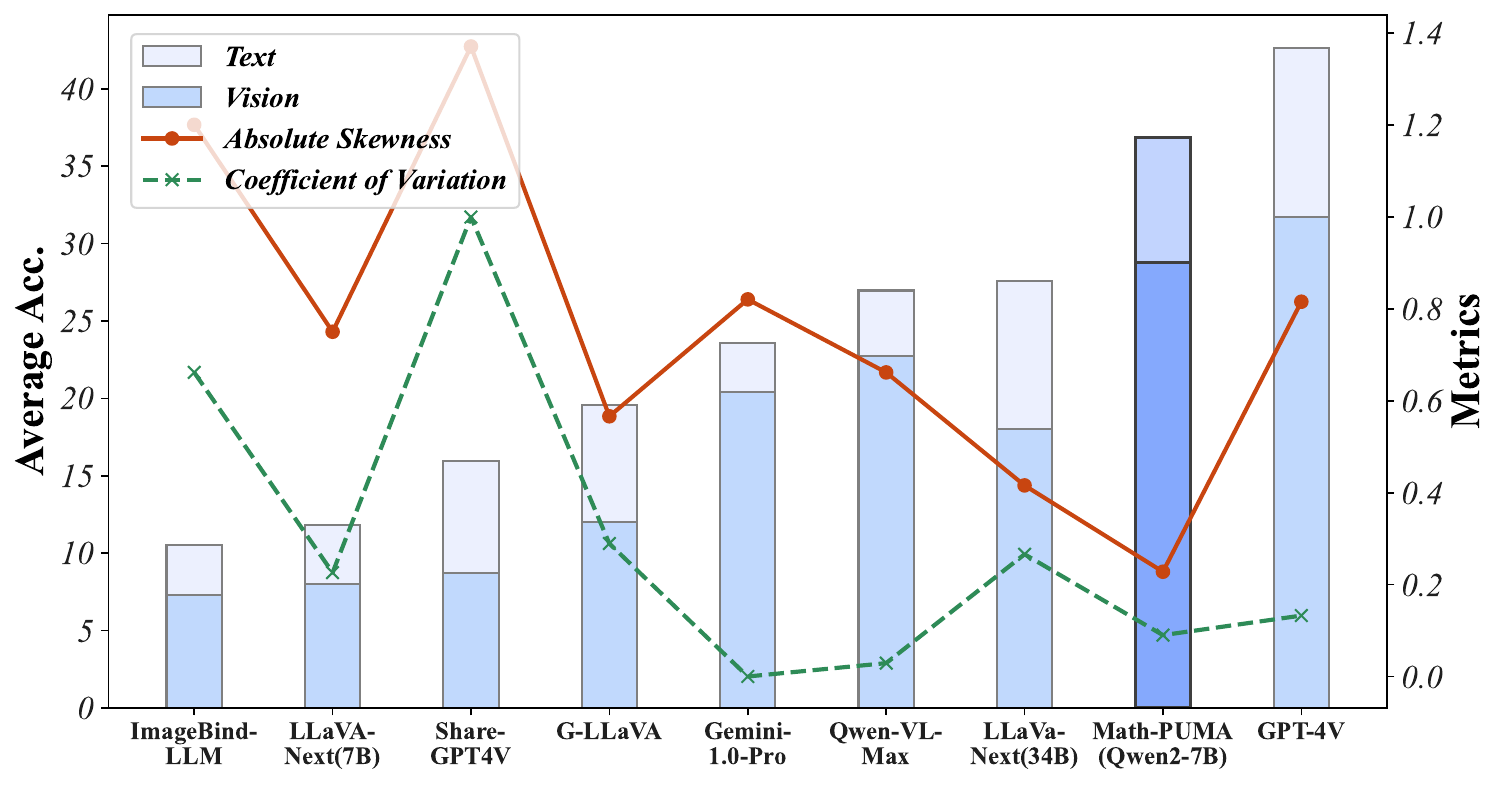}
		\caption{Visualizing MLLMs' Performance on \textsc{MathVerse}. ``Text" represents the average scores for text-dominant and text-lite categories, while ``Vision" represents the average scores for vision-intensive, vision-dominant, and vision-only categories. ``Absolute Skewness" and ``Coefficient of Variance" denote the statistical measures of score distribution across the five categories, with skewness taken as an absolute value.} \label{fig:metrics}
	\end{minipage}
\end{figure}

\subsubsection{Have the modality gaps truly narrowed?}
Through the aforementioned analysis, we have demonstrated the efficacy of our method. However, we still seek to provide a definitive conclusion to address the initial query: Has the performance gap between different modalities truly narrowed? To this end, we base our exploration on the evaluation metrics provided by \textsc{MathVerse}, calculating the average scores of the model on text-based questions and visual questions to intuitively assess the model's performance across these two distinct modalities. Additionally, we compute the skewness and coefficient of variation of the model's scores on different types of questions in \textsc{MathVerse} to corroborate our observations regarding the model's modal balance.

As illustrated in Figure \ref{fig:metrics}, we compare our model, trained using our proposed method, with several popular MLLMs.
In terms of overall performance, our model attains high average scores on both text and image-based questions, outperforming closed-source MLLMs such as Gemini-1.0-Pro and Qwen-VL-Max.
We analyze the performance gap between text and visual modalities. Our model maintains a high level of performance while exhibiting a relatively smaller gap, which is even less than that of GPT-4V.
Additionally, regarding score distribution, a model that performs consistently across modalities should demonstrate similar scores across various types of questions in \textsc{MathVerse}. Such consistency is indicated by lower absolute skewness and coefficient of variation. By visualizing the score distributions of multiple models, it is evident that our model exhibits low levels of both skewness and coefficient of variation, indicating a well-balanced performance across different types. In summary, our alignment method effectively mitigates the performance disparity between different modalities.

\section{Conclusion}
In this paper, we present Math-PUMA, a progressive upward multimodal alignment approach aimed at enhancing the mathematical reasoning capabilities of MLLMs. Experimental results indicate that Math-PUMA MLLMs not only achieve state-of-the-art performance among open-source models on multiple mathematical benchmarks but also significantly reduce the performance gap between textual and visual modalities.
Despite the impressive results of Math-PUMA, a undeniable disparity remains when compared to human-level proficiency. Continued exploration in high-quality data augmentation, automated data generation methods, and effective training strategies is necessary to further advance the mathematical reasoning abilities of MLLMs. We hope our work provides valuable insights and inspiration for future research in this domain.

\bibliography{main}

\appendix

\section{Details of Automatic Data Generation}
We provide implementation details of automatic data generation for three major types of mathematical problems as follows:
\begin{itemize}
	\item \textbf{Plane Geometry:} Triangles, quadrilaterals, and circles are the primary elements for plane geometry. Taking triangles as an example. Initially, the Question Designer constructs problems using the three basic known conditions as premises. Subsequently, the Solver for triangles utilizes these known conditions to determine all the other properties of the triangle. The Solver leverages sine and cosine rules for calculating all the properties of a triangle, including sides, angles, perimeter, and area, by providing at least three basic known conditions: side-side-side (SSS), side-angle-side (SAS), angle-side-angle (ASA), or angle-angle-side (AAS).  Moreover, the Solver enriches the problem-solving process through GPT-4o-mini. Finally, the Plotter plots the triangle and sends it to the Pair Constructor for assembly into a data pair.
	\item \textbf{Solid Geometry:} We focus on five distinctive solid geometric shapes, including cylinders, cones, prisms, pyramids, and spheres. Random values for the base radius and height are assigned for cylinders and cones, while for prisms and pyramids, random values for the length of base sides and height are given. As for spheres, only the radius is randomized.
	\item \textbf{Functions:} We focus on six primary function types, i.e., polynomial, absolute value, logarithmic, sine, cosine, and tangent functions. Initially, we develop a tool for the Plotter that enables the plotting of function graphs based on user input of the function type and parameters. Next, the Question Designer generate a variety of functions by randomly selecting the function type and parameters. Subsequently, the Question Designer utilize specific details to formulate problem statements while reserving other information for the problem-solving process. For example, given that a function passes through multiple points, we aim to determine the function's expression and properties such as zeros, symmetry axis, and period.
\end{itemize}

\section{Training Datasets Details}
A diverse and extensive dataset is the most crucial element of MLLM training. Our dataset can be divided into three parts:  200K text-only supervised fine-tuning data for stage 1, 692K alignment data pairs for stage 2 and 996K multimodal intruction tuning data for stage 3.
\subsection{Stage 1}
We extract 200K data from the existing four datasets for supervised fine-tuning of the LLM. The detailed descriptions of the four sources are as follows:
\begin{itemize}
	\item \textbf{DART-Math-Hard} \cite{tong2024dart} comprises approximately 585K mathematical question-answer pairs, generated by using DARS-Prop2Diff on the query sets from the MATH and GSK8K training datasets. This dataset attains state-of-the-art performance on several demanding mathematical reasoning benchmarks. Unlike standard rejection sampling, it intentionally emphasizes difficult queries.
	
	\item \textbf{Orca-Math-dataset} \cite{mitra2024orca} is a high-quality synthetic dataset comprising 200K math problems. This dataset was meticulously crafted using an innovative multi-agent setup, known as Agent-Instruct, where agents work collaboratively to generate and enhance the data. Each problem is paired with solutions generated by Azure GPT-4 Turbo, ensuring both accuracy and comprehensiveness.
	
	\item \textbf{NuminaMath-CoT} \cite{numina_math_datasets} includes approximately 860K math problems, each accompanied by a Chain of Thought solution. This dataset spans a spectrum of difficulties, from Chinese high school exercises to international mathematics olympiad challenges. The NuminaMath-CoT dataset aims to enhance the mathematical reasoning prowess of LLMs through the CoT approach.
	
	\item \textbf{MathInstruct} \cite{yue2023mammoth} is a meticulously curated instruction tuning dataset designed to be both lightweight and highly generalizable. Compiled from 13 different math rationale datasets, including six newly curated specifically for this work, MathInstruct stands out for its unique focus on the hybrid use of chain-of-thought and program-of-thought rationales, with 262K math problems. This approach ensures extensive coverage across a wide range of mathematical fields, making it an invaluable resource for diverse mathematical problem-solving and instruction tuning.
\end{itemize}

\subsection{Stage 2}
\begin{figure*}[h]
	\centering
	\begin{minipage}[t]{\linewidth}
		\centering
		\includegraphics[width=1.0\linewidth]{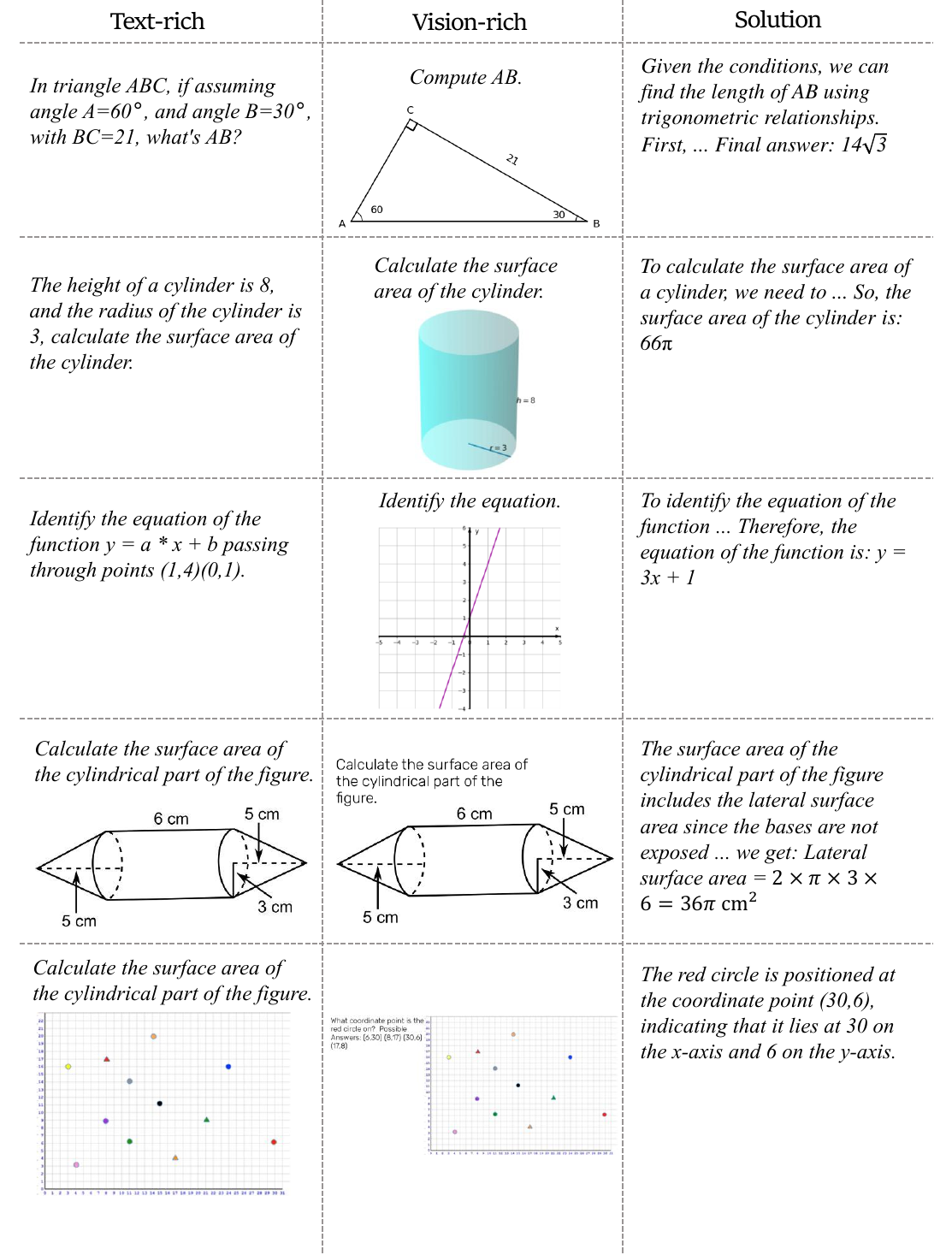}
		\caption{Examples of training dataset at stage 2.} \label{fig:traindatas2}
	\end{minipage}
\end{figure*}

We construct 692K data pairs, each pair conveying identical information but differing in multimodal representation to enhance multimodal alignment. The methods we employ to construct data pairs include automatic data generation and data augmentation based on publicly available sources. Figure \ref{fig:traindatas2} illustrates a few examples.
\begin{itemize}
	\item \textbf{Automatically Generated Dataset} includes 120K data pairs using an automated data generation pipeline, with multiple agents working collaboratively. The dataset encompasses three major categories: planar geometry, solid geometry, and functions.
	\item \textbf{Manually collecting} 80K data from online sources and augmented it to 310K using question rephrasing and image transformation methods. To streamline the creation of data pairs, we utilize a simple text-to-image rendering technique to convert the content from textual to visual form. The initial data act as the text-rich component, while the produced data form the vision-rich component.
	\item \textbf{VisualWebInstruct} dataset \cite{visualwebinstruct_datasets} comprises 262K data with science images. Corresponding question-answer pairs are generated with large language models like GPT-4o, GPT-4v, Gemini for each image. We use the same method to construct the data pairs.
\end{itemize}

\subsection{Stage 3}
\begin{figure*}[h]
	\centering
	\begin{minipage}[t]{\linewidth}
		\centering
		\includegraphics[width=1.0\linewidth]{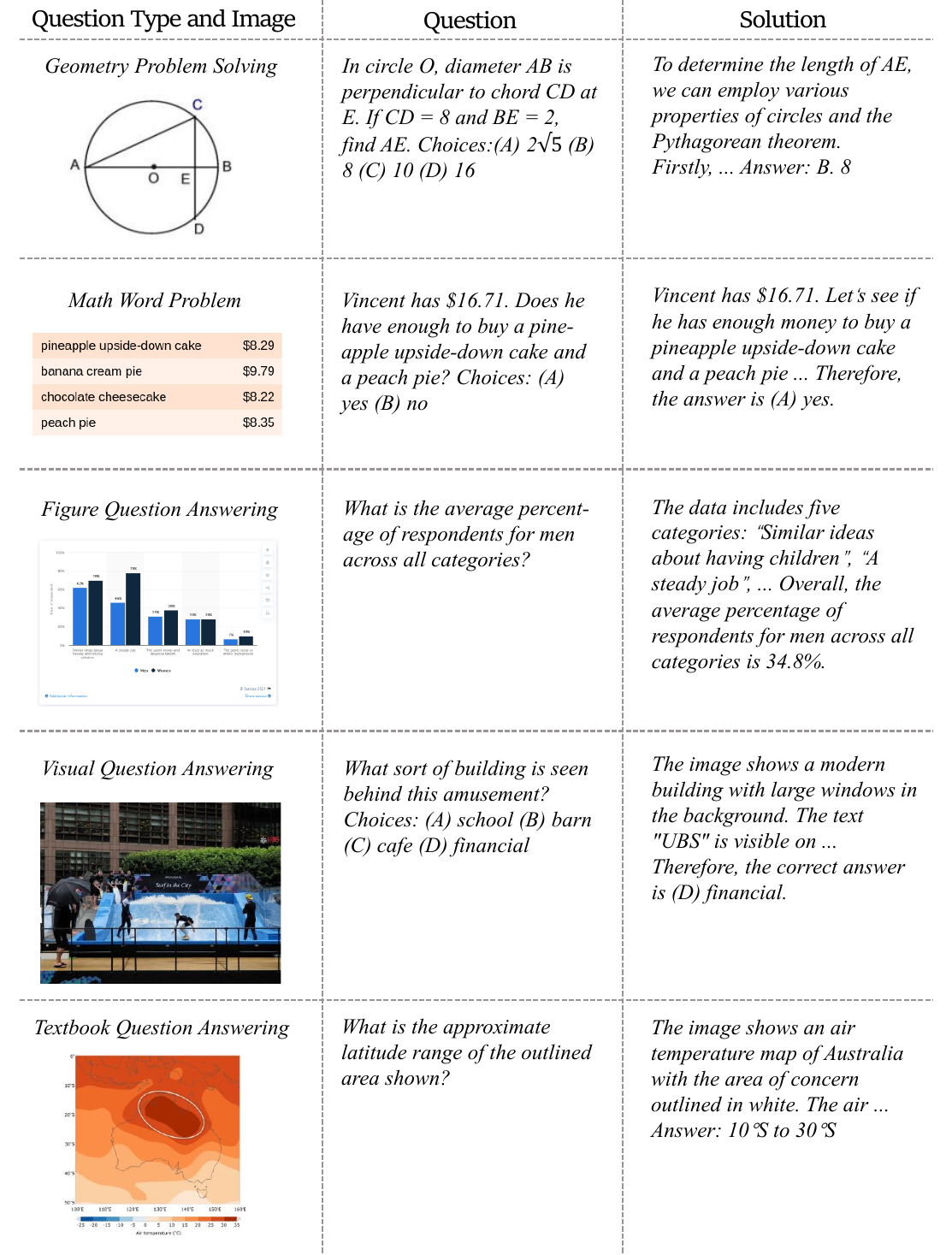}
		\caption{Examples of training dataset at stage 3.} \label{fig:traindatas3}
	\end{minipage}
\end{figure*}

We select 996K high-quality multimodal problem-solving data to fine-tune the model, further enhancing its performance in multimodal mathematical problem-solving tasks. The data sources are as follows. Figure \ref{fig:traindatas3} shows some examples.

\begin{itemize}
	\item \textbf{MathV360K} \cite{shi2024math} consists of 40K images from 24 different datasets and includes 360K question-answer pairs. It is specifically designed to enhance the multimodal mathematical reasoning capabilities of MLLMs. We enrich the geometric problem subset within MathV360K, expanding it from 40K to 120K to address the scarcity of geometric data.
	\item \textbf{VisualWebInstruct}. Using the same source as Stage 2, only the original data is used here, not the data pairs.
	\item \textbf{Manually collected dataset}. Using the same source as Stage 2, only the original data is used here, not the data pairs.
	
\end{itemize}

\section{Implementation Details}
\subsection{Enviroment}
The experiment utilizes the following libraries and their respective versions: torch=2.1.0, CUDA\_version=12.1, transfomrers=4.42.0, datasets=2.18.0, tqdm=4.40.0, Pillow=9.3.0, loguru=0.7.0.

\subsection{Hardware Configurations}
The experiments are conducted using 32 NVIDIA A100 GPUs with 80GB memory each. 
\subsection{Hyperparameters Details}
The training process is divided into 3 stages, each with specific settings tailored to optimize performance. The detailed hyperparameters of all stages are summarized in Table \ref{tab:hyper}.

\begin{table}[ht!]
	\caption{Detailed hyperparameters of our models.}
	\centering
	\small
	\begin{tabular}{l|ccc}
		\toprule
		\textbf{Hyperparameters} & \textbf{Stage 1} & \textbf{Stage 2}         & \textbf{Stage 3} \\
		\midrule
		Learning rate & $3\times10^{-5}$   & $5\times10^{-5}$ & $3\times10^{-5}$  \\
		LR scheduler  & Cosine & Cosine & Cosine  \\
		Weight decay  & 0.1 & 0.05 & 0.1   \\
		Gradient clip & 1.0 & 1.0 & 1.0  \\
		Optimizer     & \multicolumn{3}{c}{AdamW ($\beta_1=0.9, \beta_2=0.999$)}\\
		Warm-up ratio    & 0.02      & 0.02  & 0.02   \\
		raining data quantity   & 200K    & 692K & 996K \\
		Batch size       & 256      & 512 & 256   \\
		Sequence length  & 2048      & 2048 & 2048  \\
		Mixed precision training & BF16 & BF16 & BF16\\
		\bottomrule
	\end{tabular}
	\label{tab:hyper}
\end{table}

\section{Competitors Details}
In this section, we provide a comprehensive overview of the various models that serve as competitors in our comparative analysis. These models are categorized into three main groups: closed-source LLMs, closed-source MLLMs, and open-source MLLMs. 

\subsection{Closed-source LLMs}
\begin{itemize}
	\item \textbf{ChatGPT}~\cite{ouyang2022training} is developed by OpenAI. ChatGPT is a highly versatile conversational agent based on the GPT-3 architecture. 
	\item \textbf{GPT-4}~\cite{OpenAI2023GPT4TR} is also developed by OpenAI. GPT-4 is the successor to GPT-3, offering significant improvements in terms of language understanding and generation capabilities. 
\end{itemize}

\subsection{Closed-source MLLM}
\begin{itemize}
	\item \textbf{Qwen-VL-Plus}~\cite{bai2023qwen} and \textbf{Qwen-VL-Max}~\cite{bai2023qwen}: The Qwen-VL series introduces a new visual receptor, which incorporates a language-aligned visual encoder and a position-aware adapter, significantly enhancing its ability to perceive and understand visual inputs. The Qwen-VL-Plus and Qwen-VL-Max, as part of this series, stand as the enhanced and the most capable large visual language model, respectively.
	\item \textbf{Gemini-1.0-Pro}~\cite{team2023gemini} is developed by Google. It is designed to handle a wide range of tasks with impressive capabilities. Gemini-1.0-Pro is a member of the Gemini family. The Pro version is particularly balanced in terms of performance and efficiency, making it suitable for a broad spectrum of uses. 
	\item \textbf{GPT-4V}~\cite{openai2023gpt4v} is the multimodal extension of GPT-4 by OpenAI, capable of processing both text and visual data.
\end{itemize}

\begin{figure}[h]
	\centering
	\begin{minipage}[t]{\linewidth}
		\centering
		\includegraphics[width=1.0\linewidth]{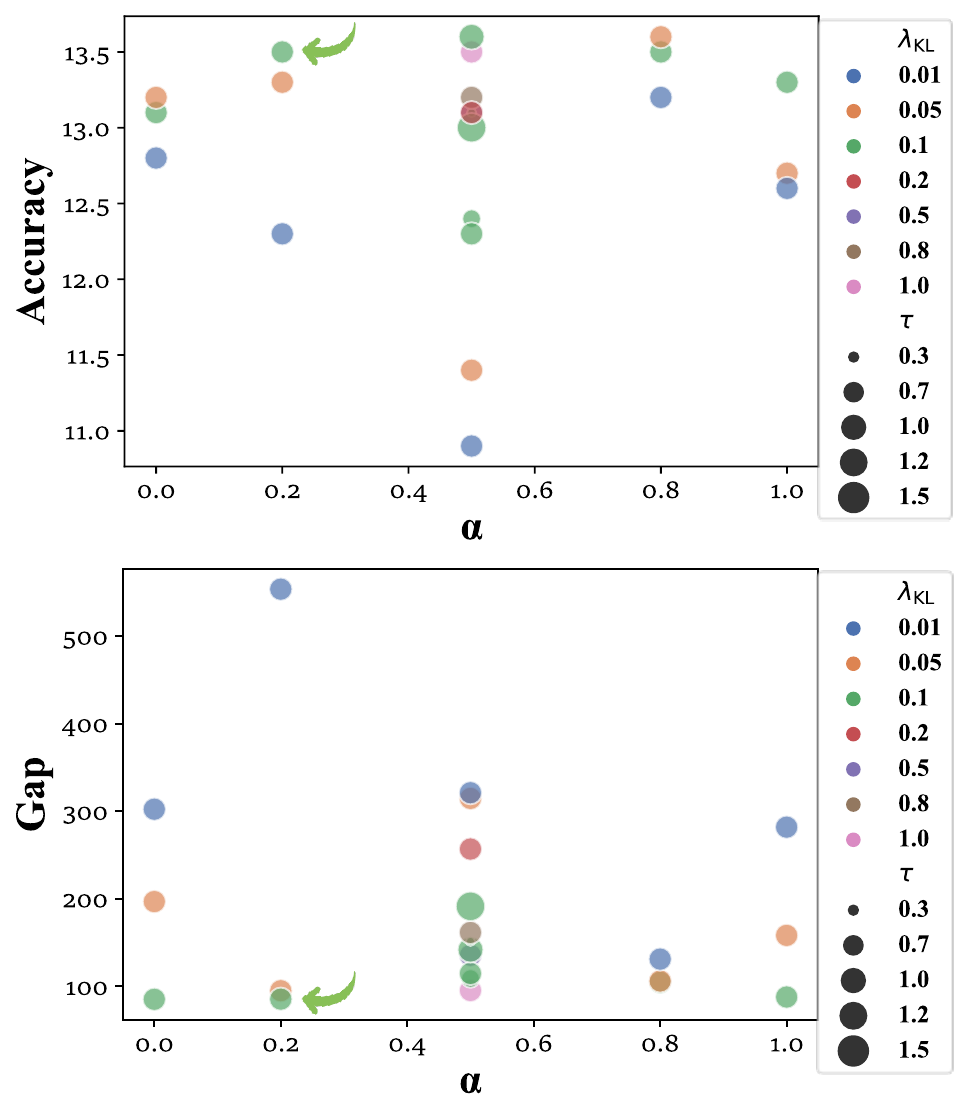}
		\caption{Visualizing hyperparameter choices, accuracy scores, and gaps on \textsc{MathVerse}. The X-axis represents $\alpha$, while the Y-axes of the top and bottom sub-figures show accuracy scores and gaps, respectively. The bubble color indicates $\lambda_\text{KL}$, and the bubble size corresponds to $\tau$. The green arrow represents our final hyperparameter choice.} \label{fig:bubble}
	\end{minipage}
\end{figure}

\subsection{Open-source MLLMs}
\begin{itemize}
	\item \textbf{mPLUG-Owl2}~\cite{ye2024mplug} features a modularized network design with a language decoder that acts as a universal interface for managing different modalities, incorporating shared functional modules and a modality-adaptive module to facilitate cross-modality interaction while preserving modality-specific features. 
	\item \textbf{LLaMA-Adapter-V2}~\cite{gao2023llamaadapterv2} is an upgrade to LLMs, enabling them to process visual data alongside text efficiently. It introduces new learnable parameters, fuses visual information early in the model, and uses a specialized joint training technique. The model leverages expert systems for enhanced image comprehension, all with minimal additional training costs.
	\item \textbf{LLaVA-1.5}~\cite{liu2024improved} and \textbf{LLaVA-NeXT}~\cite{liu2024llavanext}: LLaVA series utilize a simple mlp to connect the vision encoder and LLM. LLaVA-1.5 is the improved version of the LLaVA model, designed for better performance in multimodal tasks. LLaVA-NeXT is the latest iteration in the LLaVA series, offering enhanced multimodal capabilities and improved performance.
	\item \textbf{MiniGPT-v2}~\cite{chen2023minigpt} aims to serve as a unified interface for a variety of vision-language tasks. It utilizes unique identifiers for different tasks during training, enhancing the model's ability to distinguish and efficiently learn each task. 
	\item \textbf{SPHINX-Plus}~\cite{gao2024sphinx} is a member of the SPHINX-X family, leverages the robust LLaMA2-13B base language model. It is meticulously engineered to amplify visual and linguistic comprehension. 
	\item \textbf{ShareGPT4V}~\cite{Chen2023ShareGPT4VIL} is a novel large-scale multi-modal model that significantly enhances the alignment of vision and language modalities by leveraging a dataset of 1.2M highly descriptive image captions. These captions are generated using advanced GPT4-Vision technology, resulting in detailed and accurate descriptions that cover a wide range of information
	\item \textbf{InternLM-XComposer2}~\cite{dong2024internlm} effortlessly integrates sophisticated comprehension and creation of text-image content, transforming the way we interact with vision-language and providing fresh perspectives and possibilities.
	\item \textbf{G-LLaVA}~\cite{gllava} leverages a dataset called Geo170K, which contains over 170K meticulously crafted  geometric image-caption and question-answer pairs, to significantly enhance LLaVA's ability to interpret geometric figures and solve related problems.
	\item \textbf{SPHINX-MoE}~\cite{gao2024sphinx} is a prominent member of the SPHINX-X family. It integrates a sparse Mixture-of-Experts (MoE) architecture with 8 experts per layer, allowing it to efficiently scale up to a large parameter size without a proportional increase in computational costs. 
	\item \textbf{DeepSeek-VL}~\cite{lu2024deepseek} utilizes a hybrid vision encoder, specifically engineered to manage high-resolution images within a predetermined token limit, ensuring the model's proficiency in discerning vital semantic and complex details across a range of visual tasks.
	\item \textbf{Math-LLaVA}~\cite{shi2024math} is an enhanced version of the LLaVA-1.5 model, fine-tuned with a novel dataset called MathV360K. This dataset consists of 40K high-quality images with question-answer pairs and an additional 320K synthesized pairs to improve multimodal mathematical reasoning capabilities. 
\end{itemize}


\section{Sensitivity Analysis of Hyperparameters}
We use different combinations of hyperparameters to train MLLMs, then evaluate them on \textsc{MathVerse}. The loss is computed as
\begin{equation}
	\resizebox{0.9\linewidth}{!}
	{$
		\mathcal{L}=\lambda_{\text{KL}} (\alpha_\text{KL} \mathcal{L}_\text{FKL}+(1-\alpha_\text{KL})\mathcal{L}_\text{RKL}){\tau}^2+(1-\lambda_\text{KL})\mathcal{L}_\text{hard},
		$}
\end{equation}
where \(\lambda_{\text{KL}}\) is a hyperparameter that balances the weight between the combined FKL and RKL and the hard loss term, \(\alpha_\text{KL}\) is a weight hyperparameter that balances the contribution between \(\mathcal{L}_\text{FKL}\) and \(\mathcal{L}_\text{RKL}\).

Figure \ref{fig:bubble} shows that average accuracy scores are higher when $\lambda_\text{KL}=0.1$. Although the highest accuracy is achieved with $\lambda_\text{KL}=0.1$, $\alpha=0.5$, and $\tau=1.0$, the gap remains larger than in the setting with $\lambda_\text{KL}=0.1$, $\alpha=0.2$, and $\tau=1.0$. To balance maximizing accuracy and minimizing the gap, we ultimately select the combination $\lambda_\text{KL}=0.1$, $\alpha=0.2$, and $\tau=1.0$.

\begin{table*}[!t] \centering
	\caption{\textbf{Mathematical evaluation on \textsc{MathVerse} \textit{testmini} set.} We calculate the ``ALL" score without averaging the ``Text-only" version. For closed-source and open-source MLLMs, the best accuracy scores are marked in \textbf{bold} fonts, while the second best accuracy scores are marked in \underline{underline} fonts, respectively.}
	\setlength{\tabcolsep}{2pt}
	\resizebox{1.0\linewidth}{!}
	{
		\begin{tabular}{l|c|C{1.7cm}|C{1.7cm}|C{1.7cm}|C{1.7cm}|C{1.7cm}|C{1.7cm}|C{1.7cm}}
			\toprule
			\multirow{3}*{\large Model} & \multirow{3}*{\large \# Params.} & \multicolumn{6}{c}{\textsc{MathVerse}} \\
			\cmidrule{3-9}
			& &\multicolumn{1}{c|}{\makecell*[c]{ALL}}
			&\multicolumn{1}{c|}{\makecell*[c]{\shortstack{Text-dom.}}}
			&\multicolumn{1}{c|}{\makecell*[c]{\shortstack{Text-lite}}}
			&\multicolumn{1}{c|}{\makecell*[c]{\shortstack{Text-only}}}
			&\multicolumn{1}{c|}{\makecell*[c]{\shortstack{Vision-int.}}}
			&\multicolumn{1}{c|}{\makecell*[c]{\shortstack{Vision-dom.}}}
			&\multicolumn{1}{c}{\makecell*[c]{\shortstack{Vision-only}}}
			\\
			\midrule
			\multicolumn{9}{c}{\textit{Baselines}}\\
			\cmidrule{1-9}
			Random chance & - & 12.4 & 12.4 & 12.4 & 12.4 & 12.4 & 12.4 & 12.4 \\
			Human performance & - & 64.9 & 71.2 & 70.9 & 41.7 & 61.4 & 68.3 & 66.7 \\
			\cmidrule{1-9}
			\multicolumn{9}{c}{\textit{Closed-source LLMs}}\\
			\cmidrule{1-9}
			ChatGPT~\cite{ouyang2022training} & - & - & 33.3 & 18.9 & 33.3 & - & - & - \\
			GPT-4~\cite{OpenAI2023GPT4TR} & - & - & 46.5 & 20.7 & 46.5 & - & - & - \\
			\cmidrule{1-9}
			\multicolumn{9}{c}{\textit{Closed-source MLLMs}}\\
			\cmidrule{1-9}
			Qwen-VL-Plus~\cite{bai2023qwen} & - & 11.8 & 15.7 & 11.1 & 14.5 & 9.0 & 13.0 & 10.0 \\
			Gemini-1.0-Pro~\cite{team2023gemini} & - & 22.3 & 27.6 & 23.7 & 27.9 & 19.4 & 20.3 & 20.5 \\
			Qwen-VL-Max~\cite{bai2023qwen} & - & 24.8 & 30.3 & 24.8 & 32.2 & 20.6 & 23.3 & 25.1 \\
			GPT-4V~\cite{openai2023gpt4v} & - & \textbf{38.3} & \textbf{52.1} & \textbf{40.9} & \textbf{46.1} & \textbf{34.9} & \textbf{33.6} & \textbf{29.8} \\
			\cmidrule{1-9}
			\multicolumn{9}{c}{\textit{Open-source MLLMs}}\\
			\cmidrule{1-9}
			mPLUG-Owl2~\cite{ye2024mplug} & 7B & 4.6 & 6.6 & 6.3 & 6.1 & 6.3 & 5.6 & 4.9 \\
			LLaMA-Adapter-V2~\cite{gao2023llamaadapterv2} & 7B & 5.7 & 6.2 & 5.9 & 2.7 & 6.1 & 4.2 & 6.1 \\
			LLaVA-1.5~\cite{liu2024improved} & 13B & 7.6 & 8.8 & 7.6 & 11.5 & 7.4 & 7.4 & 6.9 \\
			LLaVA-NeXT~\cite{liu2024llavanext} & 8B & 10.3 & 12.8 & 12.0 & 9.9 & 10.7 & 9.7 & 6.3 \\
			MiniGPT-v2~\cite{chen2023minigpt} & 7B & 11.0 & 12.1 & 12.0 & 11.7 & 13.1 & 10.3 & 7.4 \\
			SPHINX-Plus~\cite{gao2024sphinx} & 13B & 12.2 & 13.9 & 11.6 & 14.9 & 11.6 & 13.5 & 10.4 \\
			ShareGPT4V~\cite{Chen2023ShareGPT4VIL} & 13B & 13.1 & 16.2 & 6.6 & 16.2 & 15.5 & 13.8 & 3.7 \\
			InternLM-XC2.~\cite{dong2024internlm} & 7B & 16.3 & 20.2 & 14.3 & 24.5 & 14.2 & 17.5 & 15.2 \\
			G-LLaVA~\cite{gllava} & 7B & 16.6 & 20.9 & 20.7 & 21.1 & 17.2 & 14.6 & 9.4 \\
			SPHINX-MoE~\cite{gao2024sphinx} & 8$\times$7B & 16.8 & 26.2 & 17.4 & 26.7 & 16.7 & 12.5 & 11.1 \\
			DeepSeek-VL~\cite{lu2024deepseek} & 7B & 19.3 & 23.0 & 23.2 & 23.1 & 20.2 & 18.4 & 11.8 \\
			Math-LLaVA~\cite{shi2024math} & 13B & 22.9 & 27.3 & 24.9 & 27.0 & 24.5 & 21.7 & 16.1 \\
			\midrule
			Math-PUMA-Qwen2-1.5B & 1.5B & 29.6 & 35.8 & 32.2 & 35.2 & 31.3 & \underline{30.4} & \underline{18.5} \\
			Math-PUMA-Qwen2-7B & 7B & \textbf{33.6} & \underline{42.1} & \underline{35.0} & \underline{39.8} & \underline{33.4} & \textbf{31.6} & \textbf{26.0} \\
			Math-PUMA-DeepSeek-Math-7B & 7B & \underline{31.8} & \textbf{43.4} & \textbf{35.4} & \textbf{47.5} & \textbf{33.6} & \textbf{31.6} & 14.7 \\
			\bottomrule
		\end{tabular}
	}
	\label{tab:mathverse_text_only}
\end{table*}

\section{Detailed Experiment Results}
Our primary evaluation was conducted on three benchmarks: \textsc{MathVerse}, \textsc{MathVista}, and \textsc{We-Math}. The experimental results are discussed in detail below.

\subsection{Performance on \textsc{MathVerse}}
The distinguishing feature of \textsc{MathVerse} is its capacity to evaluate the model's performance on identical information problems under varying degrees of multimodal richness. Consequently, an examination of the discrepancy between each category represents an intriguing and valuable undertaking. Before the discussion, let us briefly introduce the six evaluation categories of \textsc{MathVerse}:

\begin{table*}[!t]
	\centering
	\caption{\textbf{Mathematical Evaluation on different subjects and subfields in \textsc{MathVerse} \textit{testmini} set.} ‘TO’ indicates that the scores are averaged over five problem versions of the 'Text-only' version. Without TO, the scores are averaged over five problem versions, excluding the ‘Text-only’ version.  Len: Length; Anal: Analytic; Apply: Applied; Vol: Volume; Coord: Coordinate; Prop: Property; Exp: Expression; Apply: Applied. The best accuracy scores without TO are marked in \textbf{bold} fonts.}
	\setlength{\tabcolsep}{2pt}
	\resizebox{1.0\linewidth}{!}{
		\begin{tabular}{l|c|C{0.9cm}C{0.9cm}C{0.9cm}C{0.9cm}C{0.9cm}C{0.9cm}|C{0.9cm}C{0.9cm}C{0.9cm}C{0.9cm}|C{0.9cm}C{0.9cm}C{0.9cm}C{0.9cm}C{0.9cm}}
			\toprule
			\multirow{3}*{\makecell*[l]{\large Model}}    &\multirow{3}*{\makecell*[c]{\# Params.}}
			&\multicolumn{6}{c|}{\makecell*[c]{\shortstack{Plane Geometry}}} 
			&\multicolumn{4}{c|}{\makecell*[c]{\shortstack{Solid Geometry}}}
			&\multicolumn{5}{c}{\makecell*[c]{\shortstack{Functions}}}\\
			\cmidrule{3-17}
			& &\makecell*[c]{All} &\makecell*[c]{Len} &\makecell*[c]{Area} &\makecell*[c]{Angle} &\makecell*[c]{Anal} &\makecell*[c]{Apply} &\makecell*[c]{All} &\makecell*[c]{Len} &\makecell*[c]{Area} &\makecell*[c]{Vol} &\makecell*[c]{All} &\makecell*[c]{Coord} &\makecell*[c]{Prop} &\makecell*[c]{Exp} &\makecell*[c]{Apply} \\
			\midrule
			LLaVA-1.5 & 7B & 15.9 & 15.7 & 14.5 & 18.8 & 7.0 & 14.8 & 2.7 & 4.2 & 1.8 & 2.7 & 20.9 & 7.5 & 26.8 & 8.1 & 26.0  \\
			LLaVA-1.5 (TO) & 7B & 16.3 & 15.2 & 17.0 & 15.5 & 16.3 & 20.3 & 5.9 & 4.2 & 0.0 & 11.8 & 20.1 & 18.8 & 21.1 & 6.2 & 30.0  \\
			Math-LLaVA & 13B & 27.6 & 28.1 & 29.8 & 31.6 & 7.9 & 26.4 & 3.5 & 7.5 & 1.8 & 3.1 & 22.1 & 8.7 & 22.3 & 8.1 & 38.5 \\
			Math-LLaVA (TO) & 13B & 33.5 & 32.9 & 38.3 & 37.3 & 14.0 & 33.3 & 5.9 & 12.5 & 2.3 & 5.9 & 22.0 & 31.2 & 19.7 & 3.1 & 37.5 \\
			\midrule
			Math-PUMA-Qwen2-7B & 7B & 37.1 & \textbf{41.1} & \textbf{37.0} & 37.9 & \textbf{32.6} & \textbf{28.1} & \textbf{17.5} & \textbf{23.3} & \textbf{7.7} & \textbf{23.1} & \textbf{34.7} & \textbf{43.8} & \textbf{39.4} & 13.8 & \textbf{39.5} \\
			Math-PUMA-Qwen2-7B (TO) & 7B & 43.7 & 46.2 & 38.3 & 45.1 & 39.5 & 40.6 & 35.8 & 45.8 & 18.2 & 29.4 & 35.8 & 43.8 & 43.7 & 25.0 & 27.5 \\
			Math-PUMA-DeepSeek-Math-7B & 7B & \textbf{37.8} & 38.7 & 36.6 & \textbf{45.0} & 23.3 & 25.8 & 9.2 & 12.5 & 5.5 & 11.0 & 29.1 & 28.7 & 29.0 & \textbf{16.9} & 39.0 \\
			Math-PUMA-DeepSeek-Math-7B (TO) & 7B & 50.2 & 53.2 & 44.7 & 54.9 & 44.2 & 37.7 & 39.4 & 37.5 & 18.2 & 35.3 & 52.2 & 62.5 & 52.1 & 59.4 & 42.5 \\
			\bottomrule
		\end{tabular}
	}
	\label{tab:mathverse_plane_geo}
\end{table*}

\begin{itemize}
	\item \textbf{Text-dominant:} The most prevalent single-image format for text-based inquiries is one in which the entirety of the original question's textual content is retained. In most cases, the image serves as an additional clarification to the text; however, it may occasionally contain essential information.
	\item \textbf{Text-lite:} Redundant information is removed from the original question, ensuring that only essential information remains in the text, avoiding any repetition of information conveyed by both text and image. This category assesses the model's multimodal capabilities, as the absence of effective information extraction from the image would render the question incomplete.
	\item \textbf{Vision-intensive:} In comparison to text-lite, this category represents a further reduction in implicit attributes within the text. However, it is acknowledged that the transfer of information is not always complete, and vision-intensive questions may sometimes confuse humans as well.
	\item \textbf{Vision-dominant:} This category is not merely an enhancement of vision-intensive in terms of visual information but rather compared to text-lite, it annotates necessary textual information within the image, such as angle measurements, lengths of sides, etc.
	\item \textbf{Vision-only:} This category retains only the image, transferring all text to the image. Thus, it assesses not only the model's image recognition capability but also its OCR capability, as the question still appears in text format within the image.
	\item \textbf{*Text-only:} This category needs to be mentioned separately as it is not a purely textual version equivalent to the original question. Unlike the text-only concept mentioned in our main text, it merely removes the image from the text-dominant version, sometimes lacking some information such as the basic shape of functions, etc. According to \textsc{MathVerse}, this category helps determine whether MLLMs primarily rely on descriptive information or contextual visual information from diagrams to solve problems. We provide it here for reference.
\end{itemize}

As shown in Table \ref{tab:mathverse_text_only}, our model leads across multiple metrics, surpassing all similarly scaled open-source models and some closed-source models, approaching the performance of the best model, GPT-4V. Additionally, it is evident that the performance gap between vision-based and text-based questions for our model is smaller compared to GPT-4V. However, we also acknowledge that there is still a significant gap between our model's performance and that of humans. Future efforts should focus on bridging this gap and aligning modalities.

Table \ref{tab:mathverse_plane_geo} presents more detailed scores for subcategories within three main categories reported by \textsc{MathVerse}: Plane Geometry, Solid Geometry, and Functions. Each category reports both the total score and the scores for their respective subcategories. It can be observed that our model has a good understanding of planar-type questions, with very optimistic scores in Plane Geometry and Functions. In contrast, the scores for Solid Geometry are relatively average. This trend is also reflected in other open-source models, indicating that current MLLMs have weaker perceptual abilities in solid geometry. This represents a potential direction for future research advancements.

\begin{table*}[h] \centering
	\caption{\textbf{Mathematical evaluation on \textsc{MathVista} \textit{testmini} set.} Task types: FQA: figure QA, GPS: geometry problem solving, MWP: math word problem, TQA: textbook QA, VQA: visual QA. Math reasoning types: ALG: algebraic, ARI: arithmetic, GEO: geometry, LOG: logical , NUM: numeric, SCI: scientific, STA: statistical. For closed-source and open-source MLLMs, the best accuracy scores are marked in \textbf{bold} fonts, while the second best accuracy scores are marked in \underline{underline} fonts, respectively.}
	\setlength{\tabcolsep}{2pt}
	\resizebox{1.0\linewidth}{!}{
		\begin{tabular}{l|c|C{0.9cm}|C{0.9cm}|C{0.9cm}|C{0.9cm}|C{0.9cm}|C{0.9cm}|C{0.9cm}|C{0.9cm}|C{0.9cm}|C{0.9cm}|C{0.9cm}|C{0.9cm}|C{0.9cm}}
			\toprule
			{ Model} & { \# Params.} 
			&\multicolumn{1}{c|}{\makecell*[c]{ALL}}
			&\multicolumn{1}{c|}{\makecell*[c]{\shortstack{FQA}}}
			&\multicolumn{1}{c|}{\makecell*[c]{\shortstack{GPS}}}
			&\multicolumn{1}{c|}{\makecell*[c]{\shortstack{MWP}}}
			&\multicolumn{1}{c|}{\makecell*[c]{\shortstack{TQA}}}
			&\multicolumn{1}{c|}{\makecell*[c]{\shortstack{VQA}}}
			&\multicolumn{1}{c|}{\makecell*[c]{\shortstack{ALG}}}
			&\multicolumn{1}{c|}{\makecell*[c]{\shortstack{ARI}}}
			&\multicolumn{1}{c|}{\makecell*[c]{\shortstack{GEO}}}
			&\multicolumn{1}{c|}{\makecell*[c]{\shortstack{LOG}}}
			&\multicolumn{1}{c|}{\makecell*[c]{\shortstack{NUM}}}
			&\multicolumn{1}{c|}{\makecell*[c]{\shortstack{SCI}}}
			&\multicolumn{1}{c}{\makecell*[c]{\shortstack{STA}}}
			\\
			\midrule
			\multicolumn{15}{c}{\textit{Baselines}}\\
			\cmidrule{1-15}
			Random chance & - & 17.9 & 15.5 & 24.1 & 4.5 & 23.4 & 24.3 & 25.8 & 13.8 & 22.7 & 13.4 & 8.8 & 15.8 & 14.3 \\
			Human performance & - & 60.3 & 59.7 & 48.4 & 73.0 & 63.2 & 55.9 & 50.9 & 59.2 & 51.4 & 40.7 & 53.8 & 64.9 & 63.9 \\
			\cmidrule{1-15}
			\multicolumn{15}{c}{\textit{Closed-source LLMs}}\\
			\cmidrule{1-15}
			ChatGPT~\cite{ouyang2022training} & - & 33.2 & 26.0 & 31.7 & 35.5 & 48.1 & 30.2 & 32.4 & 32.3 & 33.0 & 16.2 & 17.4 & 54.9 & 36.2 \\
			GPT-4~\cite{OpenAI2023GPT4TR} & - & 33.2 & 27.9 & 31.7 & 31.2 & 51.9 & 28.5 & 33.5 & 30.9 & 32.2 & 13.5 & 12.5 & 58.2 & 37.9 \\
			\cmidrule{1-15}
			\multicolumn{15}{c}{\textit{Closed-source MLLMs}}\\
			\cmidrule{1-15}
			Qwen-VL-Plus~\cite{bai2023qwen} & - & 43.3 & \textbf{54.6} & 38.5 & 31.2 & 55.1 & 34.1 & 39.1 & 32.0 & 39.3 & 18.9 & 26.4 & 59.0 & 56.1 \\
			Gemini-1.0-Pro~\cite{team2023gemini} & - & 45.2 & 47.6 & 40.4 & 39.2 & 61.4 & \textbf{39.1} & 45.2 & 38.8 & 41.0 & 10.8 & \textbf{32.6} & 54.9 & \textbf{56.8} \\
			GPT-4V~\cite{openai2023gpt4v} & - & \textbf{49.9} & 43.1 & \textbf{50.5} & \textbf{57.5} & \textbf{65.2} & 38.0 & \textbf{53.0} & \textbf{49.0} & \textbf{51.0} & \textbf{21.6} & 20.1 & \textbf{63.1} & 55.8 \\
			\cmidrule{1-15}
			\multicolumn{15}{c}{\textit{Open-source MLLMs}}\\
			\cmidrule{1-15}
			mPLUG-Owl2~\cite{ye2024mplug} & 7B & 22.2 & 22.7 & 23.6 & 10.2 & 27.2 & 27.9 & 23.6 & 19.2 & 23.9 & 13.5 & 12.7 & 26.3 & 21.4 \\
			LLaMA-Adapter-V2~\cite{gao2023llamaadapterv2} & 7B & 23.9 & 21.2 & 25.5 & 11.3 & 32.3 & 31.8 & 26.3 & 20.4 & 24.3 & \textbf{24.3} & 13.9 & 29.5 & 18.3 \\
			LLaVA-1.5~\cite{liu2024improved} & 13B & 25.7 & 23.1 & 18.3 & 22.0 & 29.1 & \underline{39.1} & 19.6 & 28.6 & 17.6 & 10.8 & 27.8 & 33.6 & 22.9 \\
			MiniGPT-v2~\cite{chen2023minigpt} & 7B & 23.1 & 18.6 & 26.0 & 13.4 & 30.4 & 30.2 & 28.1 & 21.0 & 24.7 & 16.2 & 16.7 & 25.4 & 17.9 \\
			InternLM-XC2.~\cite{dong2024internlm} & 7B & \underline{47.8} & \textbf{53.2} & 31.7 & \textbf{76.3} & 39.2 & 36.3 & 32.0 & \textbf{51.6} & 30.5 & 13.5 & \textbf{43.8} & 37.7 & \textbf{62.8}  \\
			G-LLaVA~\cite{gllava} & 7B & 23.8 & 16.0 & 38.9 & 14.0 & 24.1 & 27.9 & 36.3 & 18.4 & 35.6 & 16.2 & 18.1 & 20.5 & 14.6  \\
			SPHINX-MoE~\cite{gao2024sphinx} & 8$\times$7B & 42.3 & \underline{49.8} & 31.2 & 42.5 & \textbf{46.8} & \textbf{39.7} & 31.7 & 41.6 & 30.5 & 16.2 & 27.1 & \textbf{50.8} & 50.8 \\
			DeepSeek-VL~\cite{lu2024deepseek} & 7B & 34.9 & 26.8 & 28.4 & 55.9 & 32.9 & 34.6 & 29.2 & 38.8 & 27.2 & 18.9 & \underline{43.1} & 35.3 & 33.2 \\
			Math-LLaVA~\cite{shi2024math} & 13B & 38.3 & 37.2 & 29.3 & 55.9 & 36.7 & 33.5 & 28.5 & 39.1 & 30.5 & 18.9 & 36.8 & 42.6 & 42.2 \\
			\midrule
			Math-PUMA-Qwen2-1.5B & 1.5B & 44.5 & 38.3 & \underline{47.6} & \underline{70.4} & 37.3 & 29.6 & \underline{43.4} & 43.9 & \textbf{47.3} & 16.2 & 34.0 & 41.0 & 47.5 \\
			Math-PUMA-Qwen2-7B & 7B & \textbf{47.9} & 46.5 & \textbf{48.1} & 68.3 & \underline{46.2} & 30.2 & \textbf{47.7} & \underline{46.2} & \textbf{47.3} & \underline{21.6} & 32.6 & 42.6 & \underline{55.8} \\
			Math-PUMA-DeepSeek-Math-7B & 7B & 44.7 & 42.8 & 39.9 & 67.7 & 42.4 & 31.3 & 39.2 & 41.9 & \underline{41.4} & 8.1 & 36.8 & \underline{48.4} & 52.5 \\
			\bottomrule
		\end{tabular}
	}
	\label{tab:mathvista}
\end{table*}

\begin{table*}[h!] \centering
	\caption{\textbf{Accuracy scores on \textsc{MathVista} \textit{testmini} set.} Abs. scene: Abstract scene; Doc. image: Document image; Func. plot: Function plot; Geo. diagram: Geometry diagram; Med. image: Medical image; Sci. figure: Scientific figure; Syn. scene: Synthetic scene. The best accuracy scores are marked in \textbf{bold} fonts.}
	\setlength{\tabcolsep}{2pt}
	\resizebox{1.0\linewidth}{!}{
		\begin{tabular}{l|c|C{1.2cm}|C{1.2cm}|C{1.2cm}|C{1.2cm}|C{1.2cm}|C{1.2cm}|C{1.2cm}|C{1.2cm}|C{1.2cm}|C{1.2cm}|C{1.2cm}|C{1.2cm}|C{1.2cm}|C{1.2cm}|C{1.2cm}}
			\toprule
			Model & \small{\# Params.} & Abs. scene & Bar chart & Doc. image & Func. plot & Geo. diagram & Line plot & Map chart & Med. image & Natural image & Pie chart & Puzzle test & Scatter plot & Sci. figure & Syn. scene & Table \\
			\cmidrule{1-17}
			LLaVA-1.5~\cite{liu2024improved} & 7B & 24.59 & 18.49 & 16.67 &	33.87 &	22.22 &	30.77 &	\textbf{62.50} & 33.33 & 22.94 & 33.33 & 16.67 & 27.78 & 29.35 & 32.26 & 18.57 \\
			LLaVA-1.5~\cite{liu2024improved} & 13B & 24.59 & 21.85 & 16.67 & 25.81 & 18.52 & 30.77 & 50.00 & 66.70 &	\textbf{30.28} & 41.67 & 11.11 & 25.00 & 30.43 & 29.52 & 15.71 \\
			InternLM-XC2.~\cite{dong2024internlm} & 7B & 78.69 &	\textbf{73.11} &	\textbf{41.67} &	37.10 &	30.56 &	43.59 &	\textbf{62.50} &	33.33 &	29.36 &	\textbf{66.67} &	13.89 &	\textbf{50.00} &	\textbf{39.13} &	\textbf{62.90} &	\textbf{70.00} \\
			G-LLaVA~\cite{gllava} & 13B & 27.87 &	10.08 &	0.00 &	33.87 &	37.96 &	38.46 &	37.50 & \textbf{100.00} & 13.76 &	16.67 & 16.67 &	16.67 &	17.39 &	28.23 &	7.14 \\
			DeepSeek-VL~\cite{lu2024deepseek} & 7B & \textbf{81.97} & 22.69 & 	16.67 &	37.10 &	27.78 &	41.03 &	\textbf{62.50} &	33.30 &	26.61 &	41.67 &	\textbf{19.44} & 25.00 &	28.26 &	43.55 &	50.00 \\
			Math-LLaVA~\cite{shi2024math} & 13B & 55.74 & 42.86 & 16.67 & 30.65 & 31.94 & 43.59 & \textbf{62.50} & 33.33 & 22.94 & 41.67 & \textbf{19.44} & 33.33 & 41.30 & 50.81 & 48.57 \\
			\midrule
			Math-PUMA-Qwen2-1.5B & 1.5B & 72.13 & 42.02 & 16.67 & 37.10 & 48.61 & \textbf{53.85} & 37.50 &	\textbf{100.00} & 15.60 & 58.33 & 16.67 & 38.89 & 36.96 & 56.45 & 64.29 \\
			Math-PUMA-Qwen2-7B & 7B & 65.57 & 54.62 & 33.33 & \textbf{56.45} & \textbf{50.00} &	51.28 &	\textbf{62.50} & 33.00 & 16.51 & 58.33 & \textbf{19.44} & 44.44 & \textbf{39.13} & 54.03 & \textbf{70.00} \\
			Math-PUMA-DeepSeek-Math-7B & 7B & 72.13 & 42.02 & 16.67 & 37.10 & 48.61 & \textbf{53.85} & 37.50 & \textbf{100.00} & 15.60 & 58.33 & 16.67 & 38.89 & 36.96 & 56.45 & 64.29 \\
			\bottomrule
		\end{tabular}
	}
	\label{tab:mathvista_details3}
\end{table*}

\begin{table*}[h!] \centering
	\caption{\textbf{Evaluation results on \textsc{MathVista} \textit{testmini} set.} The best scores of each category are marked in \textbf{bold} fonts.}
	\setlength{\tabcolsep}{2pt}
	\resizebox{1.0\linewidth}{!}{
		\begin{tabular}{l|c|C{1.4cm}|C{1.4cm}|C{1.4cm}|C{1.4cm}|C{1.4cm}|C{1.4cm}|C{1.4cm}|C{1.4cm}|C{1.4cm}|C{1.4cm}}
			\toprule
			\multirow{3}*{\large Model} & \multirow{3}*{\large \# Params.} & \multicolumn{2}{c|}{Question Type} & \multicolumn{4}{c|}{Grade} & \multicolumn{2}{c|}{Language} & \multicolumn{2}{c}{Category}\\
			\cmidrule{3-12}
			& & Free Form & Multi Choice & \small{Elementary School} & High School & College & \small{Not Applicable} & Chinese & English & General-VQA & \small{Math-Targeted-VQA} \\ 
			\midrule
			LLaVA-1.5~\cite{liu2024improved} & 7B & 9.13 &	38.89 & 15.92 & 27.12 &	25.89 & 28.35 &	29.03 &	25.00 &	30.87 & 20.37 \\
			LLaVA-1.5~\cite{liu2024improved} & 13B & 14.35 & 35.37 & 21.39 & 24.18 & 20.54 & 30.71 & 20.97 & 26.07 & 33.26 & 19.26 \\
			InternLM-XC2.~\cite{dong2024internlm} & 7B & \textbf{50.22} & 45.74 &	\textbf{63.68} & 42.16 &	22.32 &	\textbf{51.44} &	40.32 &	\textbf{48.40} &	\textbf{52.39} & 43.89 \\
			G-LLaVA~\cite{gllava} &7B&	5.43 	&39.44 &	15.92 	&33.01 &	23.21 	&20.73 &	51.61 &	22.01 	&22.17 &	25.19 \\
			DeepSeek-VL~\cite{lu2024deepseek} & 7B	&27.61& 	41.11 &	51.74 &	32.68 	&28.57 	& 29.66 &	32.26 &	35.04 &	32.17 &	37.22 \\
			Math-LLaVA~\cite{shi2024math} & 13B & 30.65 & 44.81 & 48.26 & 38.89 &	24.11 & 36.75 & 27.42 & 39.10 & 40.22 &36.67 \\
			\midrule
			Math-PUMA-Qwen2-1.5B & 1.5B	& 36.74 & 51.11 & 59.20 & 50.98 &	30.36 & 35.70 &	48.39 & 44.34 & 38.70 & 49.44 \\
			Math-PUMA-Qwen2-7B	& 7B & 42.83 & \textbf{52.22} & 57.71 & \textbf{52.29} &	\textbf{41.07} & 41.21 &	\textbf{56.45} & 47.44 &	43.26 & \textbf{51.85} \\
			Math-PUMA-DeepSeek-Math-7B & 7B & 36.52 & 51.67 & 57.21 & 47.39 & 33.93 & 39.11 &	40.32 &	45.09 & 43.48 &	45.74 \\
			\bottomrule
		\end{tabular}
	}
	\label{tab:mathvista_details2}
\end{table*}

\begin{table*}[h!] \centering
	\caption{\textbf{Accuracy scores on \textsc{We-Math} \textit{testmini} set.} S1: one-step problem, S2: two-step problem, S3: three-step problem, Mem: Measurement, PF: Plane Figures, SF: Solid Figures, TMF: Transformations and Motion of Figures, PD: Position and Direction. AL: Angles and Length, UCU: Understanding and Conversion of Units, CPF: Calculation of Plane Figures, UPF: Understanding of Plane Figures, CSF: Calculation of Solid Figures, USF: Understanding of Solid Figures, BTF: Basic Transformations of Figures, CCF: Cutting and Combining of Figures, Dir:
		Direction, Pos: Position, RoM: Route Map, CCP: Correspondence of Coordinates and Positions. For open-source MLLMs ($\approx$10B), the best scores of each category are marked in \textbf{bold} fonts.}
	\setlength{\tabcolsep}{1pt}
	\resizebox{1.0\linewidth}{!}{
		\begin{tabular}{l|c|C{0.9cm}|C{0.9cm}|C{0.9cm}|C{0.9cm}|C{0.9cm}|C{0.9cm}|C{0.9cm}|C{0.9cm}|C{0.9cm}|C{0.9cm}|C{0.9cm}|C{0.9cm}|C{0.9cm}|C{0.9cm}|C{0.9cm}}
			\toprule
			\multirow{2}{*}{Model} & \multirow{2}{*}{\# Params.} & \multirow{2}{*}{S1} & \multirow{2}{*}{S2} &\multirow{2}{*}{S3} & \multicolumn{2}{c}{Mem} & \multicolumn{2}{|c}{PF} & \multicolumn{2}{|c}{SF} & \multicolumn{2}{|c}{\centering{TMF}} & \multicolumn{4}{|c}{PD} \\
			
			\cmidrule{6-17}
			
			& & & & & UCU & AL & CPF & UPF & CSF & USF & BTF & CCF & Dir & Pos & RoM & CCP \\
			
			\cmidrule{1-17}
			\multicolumn{17}{c}{\textit{Closed-source MLLMs}}\\
			\cmidrule{1-17}        
			Qwen-VL-Max~\cite{bai2023qwen} & - & 40.8 &	30.3 & 20.6 & 19.4 & 25.3 & 39.8 & 41.4 & 43.6 & 48.0 &	43.8 & 26.7 & 41.4 & 35.1 & 40.7 & 26.7 \\
			Gemini-1.5-Pro~\cite{reid2024gemini} & - & 56.1 & 51.4 & 33.9 & 51.0 &	31.2 & 61.8 & 45.0 & 70.0 & 57.5 & 39.2 & 60.0 & 68.8 & 54.1 & 40.7 & 60.0 \\
			GPT-4V~\cite{openai2023gpt4v} & - & 65.5 & 49.2 & 38.2 & 82.5 & 38.4 & 70.7 & 60.2 & 76.6 & 56.3 & 57.8 & 63.3 & 79.3 & 57.5 & 47.8 & 63.3 \\
			GPT-4o~\cite{openai2024gpt4o} & - & 73.3 & 57.2 & 46.1 & 87.1 & 45.8 & 76.7 & 71.0 & 82.2 & 65.7 & 58.1 & 70.0 & 93.1 & 80.4 & 58.8 & 70.0 \\
			\cmidrule{1-17}
			\multicolumn{17}{c}{\textit{Open-source MLLMs ($\geq$20B)}}\\
			\cmidrule{1-17}
			InternVL-Chat-V1.5~\cite{chen2024far} & 26B & 49.4 & 30.6 & 28.5 & 44.0 & 29.8 & 52.2 & 52.1 & 44.2 & 48.2 & 47.1 & 36.7 & 65.7 & 50.5 & 36.5 & 36.7 \\
			LLaVA-NeXT~\cite{liu2024llavanext} & 72B & 42.9 & 35.6 & 30.9 & 31.7 & 25.3 & 43.3 & 42.4 & 46.1 & 41.8 & 44.2 & 36.7 & 44.3 & 39.0 & 33.0 & 36.7 \\
			LLaVA-NeXT~\cite{liu2024llavanext} & 110B & 53.7 & 36.9 & 31.5 & 39.5 & 57.7 & 59.5 & 53.1 & 52.3 & 50.2 & 54.1 & 40.0 & 54.8 & 55.9 & 40.1 & 40.0 \\
			\cmidrule{1-17}
			\multicolumn{17}{c}{\textit{Open-source MLLMs ($\approx$10B)}}\\
			\cmidrule{1-17}
			LLaVA-1.5~\cite{liu2024improved} & 7B & 30.5 & 29.7 & 29.7 & 33.7 & 28.4 & 20.0 & 38.9 & 28.6 & 35.6 & 40.4 & 32.2 & 20.7 & 21.2 & 51.7 & 43.3 \\
			LLaVA-1.5~\cite{liu2024improved} & 13B & 35.4 & 30.0 & 32.7 & 27.1 & \textbf{49.0} & 25.3 & 44.1 & 31.4 & 35.7 & 48.5 & 38.9 & 43.8 & 37.7 & 36.5 & 36.7 \\
			LLaVA-1.6~\cite{liu2024llavanext} & 7B & 23.0 & 20.8 & 15.8 & 18.5 & 20.5 & 16.9 & 29.6 & 15.6 & 18.6 & 42.7 & 26.7 & 17.6 & 43.3 & 28.9 & 26.7 \\
			LLaVA-1.6~\cite{liu2024llavanext} & 13B & 29.4 & 25.3 & 32.7 & 21.7 & 23.2 & 23.4 & 34.7 & 25.3 & 26.4 & 37.5 & 30.0 & 26.9 & 28.9 & 37.1 & 30.0 \\
			DeepSeek-VL~\cite{lu2024deepseek} & 7B & 32.6 & 26.7 & 25.5 & 16.6 & 35.1 & 27.3 & 38.0 & 24.2 & 38.7 & 50.0 & 23.3 & 24.5 & 41.0 & 51.7 & 23.3 \\
			G-LLaVA~\cite{gllava} & 13B & 32.4 & 30.6 & 32.7 & 33.3 & 29.1 & 32.0 & 37.9 & 19.6 & 33.5 & 37.1 & 40.0 & 31.2 & 33.2 & 25.6 & 40.0 \\
			Math-LLaVA~\cite{shi2024math} & 13B & 38.7 & 34.2 & 34.6 & 30.3 & 17.9 & 39.2 & 40.4 & 37.1 & 37.7 & \textbf{53.0} & 51.3 & 30.8 & 30.8 & 40.9 & \textbf{46.7}  \\
			InternLM-XC2.~\cite{dong2024internlm} & 7B & 47.0 & 33.1 & 33.3 & 31.3 & 46.5 & 47.7 & 42.6 & 51.4 & 43.9 & 41.1 & 40.0 & \textbf{65.5} & \textbf{53.9} & \textbf{55.2} & 40.0 \\
			\midrule
			Math-PUMA-Qwen2-1.5B & 1.5B & 37.0 & 31.7 & 27.3 & 25.5 & 27.9 & 37.9 & 32.5 & 41.3 & 42.7 & 36.4 & 50.1 & 27.6 & 31.5 & 18.4 & \textbf{46.7} \\
			Math-PUMA-Qwen2-7B & 7B & \textbf{53.3} & \textbf{39.4} & \textbf{36.4} & 63.5 & 42.5 & \textbf{60.2} & \textbf{45.9} & \textbf{66.2} & \textbf{48.6} & 42.3 & 53.5 & 31.2 & 37.7 & 40.4 & \textbf{46.7} \\
			Math-PUMA-DeepSeek-Math-7B & 7B & 45.6 & 38.1 & 33.9 & \textbf{69.8} & 29.1 & 45.7 & 38.6 & 51.8 & 43.6 & 41.6 & \textbf{54.6} & 44.8 & 53.8 & 29.1 & 40.0 \\
			\bottomrule
		\end{tabular}
	}
	\label{tab:wemath-acc}
\end{table*}

\begin{table*}[h!] \centering
	\caption{\textbf{Evaluation results on \textsc{We-Math} \textit{testmini} set.} AVG: average score, IK: Insufficient Knowledge, IG: Inadequate Generalization, CM: Complete Mastery, RM: Rote Memorization. The best scores of each category are marked in \textbf{bold} fonts.}
	\setlength{\tabcolsep}{1pt}
	\resizebox{1.0\linewidth}{!}{
		\begin{tabular}{l|c|C{1.2cm}|C{1.2cm}|C{1.2cm}|C{1.2cm}|C{1.2cm}|C{1.2cm}|C{1.2cm}|C{1.2cm}|C{1.2cm}|C{1.2cm}}
			\toprule
			\multirow{2}*{\large Model} & \multirow{2}*{\large \# Params.} & \multicolumn{5}{c|}{Strict} & \multicolumn{5}{c}{Loose}\\
			\cmidrule{3-12}
			& & Avg $\uparrow$ & IK $\downarrow$ & IG $\uparrow$ & CM $\uparrow$ & RM $\downarrow$ & Avg $\uparrow$ & IK $\downarrow$ & IG $\uparrow$ & CM $\uparrow$ & RM $\downarrow$\\ 
			\cmidrule{1-12}
			\multicolumn{12}{c}{\textit{Closed-source MLLMs}}\\
			\cmidrule{1-12}        
			Qwen-VL-Max~\cite{bai2023qwen} & - & 10.5 & 65.1 & 7.6 & 6.7 & 75.5 & 25.5 & 65.1 & 7.6 & 21.7 & 20.3 \\
			Gemini-1.5-Pro~\cite{reid2024gemini} & - & 26.4 & 42.9 & 11.2 & 20.8 & 54.8 & 46.0 & 42.9 & 11.2 & 40.4 & 12.0 \\
			GPT-4V~\cite{openai2023gpt4v} & - & 31.1 & 39.8 & 14.5 & 23.8 & 47.9 & 51.4 & 39.8 & 14.5 & 44.2 & 3.3 \\
			GPT-4o~\cite{openai2024gpt4o} & - & \textbf{42.9} & \textbf{31.2} & \textbf{15.2} & \textbf{35.2} & \textbf{34.2} & \textbf{60.6} & \textbf{31.2} & \textbf{15.2} & \textbf{53.0} & \textbf{1.1} \\
			\cmidrule{1-12}
			\multicolumn{12}{c}{\textit{Open-source MLLMs ($\geq$20B)}}\\
			\cmidrule{1-12}
			InternVL-Chat-V1.5~\cite{chen2024far} & 26B & 12.7 & 56.4 & 10.5 & 7.4 & 77.6 & 31.0 & 56.4 & 10.5 & 25.7 & 22.4 \\
			LLaVA-NeXT~\cite{liu2024llavanext} & 72B & 13.4 & 58.9 & 7.1 & 9.9 & 71.0 & 31.5 & 58.9 & 7.1 & 28.0 & 17.9 \\
			LLaVA-NeXT~\cite{liu2024llavanext} & 110B & \textbf{19.2} & \textbf{50.3} & \textbf{14.5} & \textbf{12.0} & \textbf{66.0} & \textbf{37.9} & \textbf{50.3} & \textbf{14.5} & \textbf{30.7} & \textbf{13.0} \\
			\cmidrule{1-12}
			\multicolumn{12}{c}{\textit{Open-source MLLMs ($\approx$10B)}}\\
			\cmidrule{1-12}
			LLaVA-1.5~\cite{liu2024improved} & 7B & 6.8 & 65.1 & 5.1 & 4.2 & 85.9 & 24.1 & 65.1 & 5.1 & 21.5 & 27.6 \\
			LLaVA-1.5~\cite{liu2024improved} & 13B & 9.3 & 65.0 & 4.2 & 7.2 & 76.5 & 24.4 & 65.0 & 4.2 & 22.3 & 27.8 \\
			LLaVA-1.6~\cite{liu2024llavanext} & 7B & 3.3 & 78.3 & 2.5 & 2.1 & 89.1 & 13.8 & 78.3 & 2.5 & 12.6 & 34.7 \\
			LLaVA-1.6~\cite{liu2024llavanext} & 13B & 5.2 & 69.1 & 3.2 & 3.6 & 86.9 & 22.0 & 69.1 & 3.2 & 20.4 & 26.2 \\
			DeepSeek-VL~\cite{lu2024deepseek} & 7B & 6.3 & 69.1 & 4.6 & 4.0 & 84.8 & 21.0 & 69.1 & 4.6 & 18.7 & 29.0 \\
			Math-LLaVA~\cite{shi2024math} & 13B & 11.1 & 62.1 & 3.6 & 9.3 & 72.8 & 31.3 & 62.1 & 3.6 & 29.5 & 13.9 \\
			InternLM-XC2.~\cite{dong2024internlm} & 7B & 12.7 & 56.4 & 10.5 & 7.4 & 77.6 & 31.0 & 56.4 & 10.5 & 25.7 & 22.4 \\
			\midrule
			Math-PUMA-Qwen2-1.5B & 1.5B & 10.4 & 63.8 & 5.9 & 7.4 & 75.5 & 26.4 & 63.8 & 5.9 & 23.4 & 22.6 \\
			Math-PUMA-Qwen2-7B & 7B & \textbf{19.2} & \textbf{47.8} & \textbf{13.7} & \textbf{12.4} & 67.8 & \textbf{41.0} & \textbf{47.8} & \textbf{13.7} & \textbf{34.1} & \textbf{11.4} \\
			Math-PUMA-DeepSeek-Math-7B & 7B & 15.6 & 56.0 & 7.2 & 12.0 & \textbf{67.4} & 35.8 & 56.0 & 7.2 & 32.2 & 12.4 \\
			\bottomrule
		\end{tabular}
	}
	\label{tab:wemath-four}
\end{table*}

\subsection{Performance on \textsc{MathVista}}
\textsc{MathVista} is a comprehensive visual environment reasoning benchmark. It is not limited to traditional exam-style questions but includes a broader range of visual reasoning scenarios such as charts, real-world photographs, and external knowledge inference. Its sources are diverse, including traditional datasets like AI2D and Geometry3K, as well as newly created ones such as IQTest and PaperQA. It encompasses both real and synthetic scenes and supports multiple languages, including Chinese and English.

We will now focus more on the scores of each subcategory to explore the model's capabilities in different directions in detail.

\subsubsection{Task \& Skills}
In this section we focus on analysing the tasks and skills of the model and the results are displayed in Table \ref{tab:mathvista}.

\begin{itemize}
	\item \textbf{Advantages:} Due to our model being trained on higher-quality mathematical reasoning data, it demonstrates strong competitiveness in some rigorously formatted exam-style questions, such as:
	\begin{itemize}
		\item GPS (Geometry Problem Solving): Our most advanced model achieved an accuracy of 48.1\%, nearly matching the human performance of 48.4\%.
		
		\item MWP (Math World Problem): Our three models achieved high performance, scoring 70.4\%, 68.3\%, and 67.7\%, respectively, closely approaching the human performance of 73.0\%.
		
		\item ALG (Algebraic): Our models secured the top-2 positions among the open-source models reported, with the most advanced model achieving an accuracy of 47.7\%, very close to the human performance of 50.9\%.
		
		\item GEO (Geometry): Our models also secured the top-2 positions among the open-source models reported, with the most advanced model achieving an accuracy of 47.3\%, not far from the human performance of 51.4\%.
	\end{itemize}
	
	These results indicate that our model exhibits excellent mathematical reasoning capabilities, leading among open-source models and closely approximating human performance, which fully demonstrates the effectiveness of our approach in mathematical reasoning.
	\item \textbf{Disadvantages:} However, our model also shows room for improvement in some areas, such as:
	\begin{itemize}
		\item FQA (Figure QA): Our most advanced model has achieved 46.5\%, which is comparable to GPT-4V's 43.1\% and Gemini-1.0-Pro's 47.6\%, but still falls short of the human performance of 59.7\%.
		
		\item LOG (Logical): Our most advanced model has reached an accuracy of 21.6\%, surpassing Qwen-VL-Plus and Gemini-1.0-Pro, and being on par with GPT-4V. However, it still has a significant gap compared to the human performance of 40.7\%.
		
		\item NUM (Numeric): Our models scored 34.0\%, 32.6\%, and 36.8\%, respectively, outperforming the best closed-source model GPT-4V's 21.6\%, but still exhibiting a considerable gap from the human performance of 53.8\%.
		
	\end{itemize}
	
	Our objective is to further align and approach human-level performance, and thus, these areas remain a focal point for improvement.
\end{itemize}

\subsubsection{Context}

In Table \ref{tab:mathvista_details3}, we present the performance of our model under different contexts. Consistent with the analysis earlier, our model shows outstanding performance in geometry-related tasks. For example, our model significantly outperforms others in function plots, geometry diagrams, and line plots. For standard visual question answering tasks, such as bar charts, map charts, and pie charts, our model also performs well. However, our model underperforms in scenarios involving natural images. Our preliminary conclusion is that this is due to the limited training data for this category, which is a potential direction for future improvement.

\subsubsection{Others}
In Table \ref{tab:mathvista_details2}, we also present the scores for other categories in \textsc{MathVista}. Several interesting observations can be noted:

\begin{itemize}
	\item Our model performs well in both free-form and multiple-choice question types, indicating that the model is not only capable of selecting options but also has some dialogue and continuation abilities. We also noticed that while all open-source models perform well on multiple-choice questions, some models show significantly lower scores in free-form questions. This may be due to the training data overly focusing on common multiple-choice questions.
	
	\item Our model excels in more challenging questions at the high school level and above. It can be observed that open-source models generally perform well on elementary school-level questions but struggle with higher difficulty questions at the high school and college levels. Our model, however, performs exceptionally well on these challenging questions, demonstrating its deeper reasoning capabilities.
	
	\item Our model shows balanced performance in both Chinese and English language questions. It is noticeable that some open-source models may prefer to solve questions in a single language. We speculate that this could be due to the base LLM's limitations, such as LLaVA models commonly being trained on English corpora, or the SFT data having biases, such as certain language data and questions being easier to collect.
\end{itemize}

\subsection{Performance on \textsc{We-Math}}

\textsc{We-Math} is a benchmark that emphasizes mathematical reasoning, making it highly relevant to our research focus. Unlike existing benchmarks, it is constructed around textbook knowledge units, decomposing problem-solving into subproblems based on knowledge concepts.

\subsubsection{Accuracy}
As shown in Table \ref{tab:wemath-acc}, \textsc{We-Math} evaluates scores for S1 (one-step), S2 (two-step), and S3 (three-step) problems, as well as multiple categories of questions.

In S1, S2, and S3, our models performed exceptionally well. Our best model not only achieved state-of-the-art results among models of similar scale but also demonstrated comparable performance to significantly larger models. Specifically, our model scored 53.3, 39.4, and 36.4 for the three steps, respectively, which are comparable to LLaVA-NeXT-110B's 53.7, 36.9, and 31.5. This indicates that our model possesses reasoning abilities comparable to large-scale MLLMs, substantiating the effectiveness of our approach.

In other categories, consistent with the findings in \textsc{MathVista}, our model excels in geometric reasoning. For example, in CPF (Calculation of Plane Figures), UPF (Understanding of Plane Figures), CSF (Calculation of Solid Figures), and USF (Understanding of Solid Figures), our best model achieved the best performance among open-source models of similar scale and performed comparably to larger open-source MLLMs. Our model also demonstrated complex understanding capabilities, achieving state-of-the-art performance among models of similar scale in CCF (Cutting and Combining of Figures) and CCP (Correspondence of Coordinates and Positions).

\subsubsection{Four Dimensional Metrics}
\textsc{We-Math} provides four dimensions of evaluation metrics: IK (Insufficient Knowledge), IG (Inadequate Generalization), CM (Complete Mastery), and RM (Rote Memorization). We briefly introduce each metric for subsequent analysis:

\begin{itemize}
	\item \textbf{Insufficient Knowledge (IK):} Part of one-step problems contain errors, and the multi-step problem is wrong. This is reasonable because the model's insufficient grasp of single knowledge concepts may lead to errors in multi-step problems.
	
	\item \textbf{Inadequate Generalization (IG):} One-step problems are all correct, but the multi-step problem is incorrect. This is also considered reasonable. While LMMs are capable of understanding individual knowledge concepts, they may struggle to generalize that knowledge to solve composite problems.
	
	\item \textbf{Complete Mastery (CM):} One-step problems are all correct, and the multi-step problem is also answered correctly. This result demonstrates that the model's results are both reliable and accurate.
	
	\item \textbf{Rote Memorization (RM):} One-step problems contain errors, but the multi-step problem is answered correctly, which contradicts human logical thinking. If a model can solve composite multi-step problems but fails to answer the one-step problems needed in the process, it raises doubts about the model's reliability.
\end{itemize}

As shown in Table \ref{tab:wemath-four}, our model not only demonstrates competitiveness in the average score but also matches the performance of large-scale MLLMs like LLaVA-NeXT-110B in IK and RM metrics, and outperforms closed-source models like Qwen-VL-Max. This indicates that our model does not merely memorize or recite questions but has begun to exhibit intrinsic reasoning capabilities, showing good generalization performance and the ability to solve extended problems.

\end{document}